%% file: neurips_2026.tex
\documentclass{article}

\usepackage[preprint]{neurips_2026}

\usepackage[utf8]{inputenc} 
\usepackage[T1]{fontenc}    
\usepackage{amsmath}
\usepackage{url}            
\usepackage{booktabs}       
\usepackage{amsfonts}       
\usepackage{nicefrac}       
\usepackage{microtype}      
\usepackage{xcolor}         

\usepackage{booktabs}
\usepackage{multirow}
\usepackage{wrapfig}
\usepackage{pifont}
\newcommand{\cmark}{\ding{51}}
\newcommand{\xmark}{\ding{55}}

\usepackage{enumitem}
\usepackage{graphicx}
\usepackage{subcaption}

\usepackage{booktabs}
\usepackage{multirow}
\usepackage{threeparttable}
\usepackage{array}
\usepackage{hyperref}       

\usepackage{xurl}   

\title{TimeSage-MT: A Multi-Turn Benchmark for Evaluating Agentic Time Series Reasoning}

\author{%
  \textbf{Yaxuan Kong}\textsuperscript{1,2}\thanks{Equal contribution.} \quad
  \textbf{Qingren Yao}\textsuperscript{3}\footnotemark[1] \quad
  \textbf{Yuqi Nie} \quad
  \textbf{Yichen Li}\textsuperscript{2} \quad
  \textbf{Yilei Shao}\textsuperscript{6} \quad
  \textbf{Stefan Zohren}\textsuperscript{1} \\ 
  \textbf{Anna Vettoruzzo}\textsuperscript{3} \quad
  \textbf{Joaquin Vanschoren}\textsuperscript{3} \quad
  \textbf{Ming Jin}\textsuperscript{4}\thanks{Corresponding author.} \quad
  \textbf{Qingsong Wen}\textsuperscript{5}\footnotemark[2] \\[6pt]
  \textsuperscript{1}University of Oxford \quad
  \textsuperscript{2}VulpiVox Intelligence \quad
  \textsuperscript{3}Eindhoven University of Technology \\
  \textsuperscript{4}Griffith University \quad
  \textsuperscript{5}Squirrel Ai Learning \quad
  \textsuperscript{6}East China Normal University \\[4pt]
  \texttt{yaxuan.kong@eng.ox.ac.uk; q.yao@tue.nl}
}

\begin{document}

\maketitle

\begin{abstract}
Time series data inform critical decisions across many real-world domains. While large language model (LLM) agents can analyze data through natural language and tools, it remains unclear whether they can conduct reliable time series analysis across multi-turn conversations. Existing benchmarks focus on single-step tasks such as forecasting and anomaly detection, overlooking practical workflows where user goals evolve, agents must build on prior analyses, and conclusions emerge from accumulated evidence. In this work, we introduce TimeSage-MT, a multi-turn benchmark for agentic time series reasoning with 240 tasks and 2,680 dialogue turns across 8 real-world domains, spanning basic exploration to decision-oriented analysis. TimeSage-MT is built through a reproducible pipeline that converts real-world time series data into multi-turn conversations with verifiable answers. It provides a unified evaluation protocol and public leaderboard for comparing time series agentic systems. To demonstrate the benchmark's utility, we evaluate frontier LLMs alongside TimeSage, a novel structured agent equipped with a comprehensive time series skill library. The results show sharp performance drops on decision-oriented tasks, driven by failures in memory, uncertainty handling, and domain-based decision making. TimeSage-MT exposes critical gaps in current agentic reasoning and provides a rigorous foundation for future development.
\end{abstract}

\section{Introduction}
\label{section_1_introduction}
Time series analysis drives critical decision-making across high-stakes domains, from financial portfolio balancing to healthcare risk assessment \cite{box2015time,hyndman2021forecasting,kong2024financial}. Time series data capture how systems evolve, when risks arise, and how actions affect future outcomes \cite{nie2023patchtst,liu2025sundial,xiao2025timefound}. In practice, however, useful time series analysis rarely ends with a single prediction \cite{kong2025unlocking,paiva2026llmtsf}. Analysts often inspect data quality, identify temporal structure, choose appropriate methods, quantify uncertainty, and translate results into actions that respect domain constraints \cite{zhang2026tfllm,weng2026temporalbench}. A financial analyst may assess volatility and regime shifts before deciding whether to rebalance a portfolio, and a grid operator may combine load forecasting, anomaly detection, and uncertainty analysis before making an operational recommendation. This workflow exposes a gap between existing time series benchmarks and the reasoning behavior expected from general-purpose analytical agents. Standard benchmarks for forecasting, anomaly detection, and classification have made important progress, but typically evaluate isolated tasks with fixed objectives and task-specific metrics \cite{godahewa2021monash,makridakis2020m4,paparrizos2022tsb}. They do not test whether an agent can maintain the dialog state, select analytical methods, chain dependent computations, adapt when assumptions fail, or communicate conclusions in a domain-based form \cite{yao2023react,xu2025toolagents,chen2025toollearning}. As LLM agents become increasingly capable of natural language interaction, code execution, tool use, planning, and self-correction, this evaluation gap becomes urgent \cite{luo2025llmagent,zhao2025agenticreasoning}. Without benchmarks that evaluate the entire analytical trajectory, it is hard to distinguish reliable time series reasoning from shallow pattern matching, hallucinated numbers, accidental tool use, or plausible but unsupported explanations \cite{yehudai2025agent_eval}.

\begin{figure*}[!h]
\begin{center}
\includegraphics[width = \linewidth]{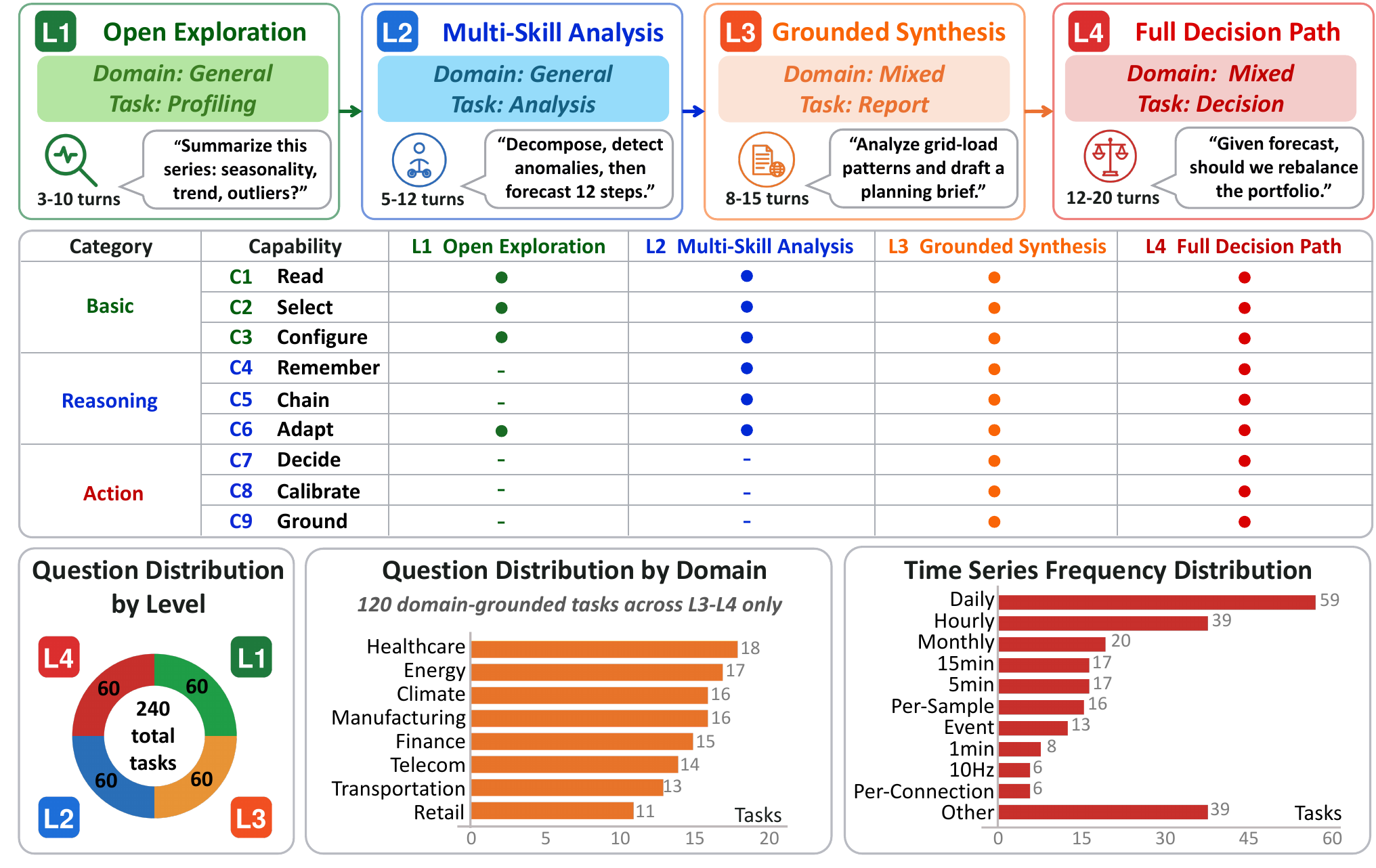}
\end{center}
\caption{Overview of TimeSage-MT. It organizes 240 tasks over 4 levels and targets 9 capabilities for agentic time series reasoning. Filled circles mark active capabilities; dashes mark inactive ones.}
\label{Figure_1}
\end{figure*}

Designing such a benchmark is challenging for three reasons. \textbf{First}, time series reasoning is inherently multi-capacity \cite{zhang2026tfllm}. As shown in Figure \ref{Figure_1}, an effective agent must read the data properties (\textit{C1}), select suitable methods (\textit{C2}), configure the analyses correctly (\textit{C3}), remember prior results across turns (\textit{C4}), chain dependent analyzes (\textit{C5}), adapt when assumptions fail or user goals change (\textit{C6}), make actionable decisions (\textit{C7}), calibrate uncertainty (\textit{C8}) and ground conclusions in domain knowledge (\textit{C9}). \textbf{Second}, the depth of reasoning varies substantially between interactions. Some conversations require only exploratory profiling and summarization, while others require long decision-oriented workflows that combine forecasting, anomaly detection, risk assessment, synthesis, and justification. \textbf{Third}, evaluation must be reproducible and interpretable. Multi-turn agent outputs can appear convincing even when the underlying analysis violates assumptions, ignores earlier turns, or relies on unsupported reasoning \cite{yehudai2025agent_eval,weng2026temporalbench}. A meaningful benchmark therefore needs diverse real-world datasets, realistic multi-turn tasks, verifiable answer targets, and a unified protocol that supports both end-to-end evaluation and fine-grained diagnosis across capabilities, domains, and task types. 

To address these challenges, we introduce \textbf{TimeSage-MT}, a multi-turn benchmark and evaluation suite for agentic time series reasoning. TimeSage-MT contains \textbf{240 tasks} and \textbf{2,680 dialog turns} across \textbf{8 real-world domains}, organized into \textbf{4 complexity levels}: open exploration, multi-skill analysis, grounded synthesis, and full decision paths. Each task is paired with analysis trajectories, reference code, numerical ground truth, and level-specific output requirements, enabling evaluation of both final answers and intermediate reasoning behavior. To ensure scalability without sacrificing rigor, we engineer a reproducible pipeline that transforms raw datasets into verifiable, multi-turn analytical trajectories. TimeSage-MT further provides a unified evaluation protocol and leaderboard to compare time series agentic systems. 
To demonstrate the benchmark's utility, we evaluate six frontier LLMs and multiple reasoning paradigms (direct, code-enabled, and skill-guided), alongside TimeSage, a novel structured agent system equipped with a comprehensive time series skill library comprising \textbf{226 analytical skills} across \textbf{36 categories}.
By comparing these settings, TimeSage-MT exposes critical gaps in current agentic reasoning from basic exploration to decision-oriented analysis. Experiments show that current systems can often handle short exploratory tasks, but performance degrades sharply when tasks require multi-turn memory, uncertainty-aware reasoning, and domain-grounded decisions. These observations reveal failures not captured by single-task benchmarks and establish TimeSage-MT as a rigorous foundation for future development.

We release the dataset at \url{https://huggingface.co/datasets/TimeSage-Series/TimeSage-MT} and the code at \url{https://github.com/TimeSage-Series/TimeSage-MT}. Our main contributions are summarized below:
\vspace{-1mm}
\begin{itemize}[leftmargin=*]
    \item \textbf{The first multi-turn benchmark for agentic time series reasoning.}
    To the best of our knowledge, TimeSage-MT is the first benchmark for evaluating agentic time series reasoning across multi-turn analytical workflows. It contains 240 tasks and 2,680 dialogue turns across 8 domains and 4 complexity levels, from open exploration to full decision paths. Each task includes analysis trajectories, reference code, numerical ground truth, and level-specific outputs, and is organized around a 9-capability taxonomy to diagnose failures beyond the accuracy of the final-answer.

    \item \textbf{A reproducible end-to-end construction and pipeline.}
    We develop an end-to-end pipeline that converts real-world time series datasets into annotated multi-turn conversations with verifiable answer targets. The pipeline covers dataset collection, conversation generation, trajectory construction, reference code generation, numerical verification, and human annotation. We also release a reusable annotation dashboard to support task review and future benchmark extension.
    
    \item \textbf{A unified evaluation protocol, reusable skill library, and key observations.}
    We provide a public leaderboard and evaluation protocol that assess agentic systems by per-turn answer accuracy and capability-level reasoning behavior. We release a reusable library of 226 analytical skills across 36 categories, an evaluation dashboard with per-turn traces and scores, and an open platform for multi-turn interaction with time series data. Under this protocol, we compare frontier LLM-based settings, including direct response, code execution, skill-guided analysis, and TimeSage, a structured agentic system built on our skill library. Our results reveal key observations that point to persistent failures in memory, uncertainty-aware reasoning, and domain-based decision-making.
\end{itemize}

\section{Related Work}
\label{sec:related}

\begin{wraptable}{r}{0.58\linewidth}
\vspace{-12pt}
\centering
\caption{Comparison with representative prior benchmarks. ``Multi'' (multi-task evaluation), ``Agent'' (tool-using evaluation), ``NL'' (natural-language interaction), ``Mix'' (checklist + judge evaluation), ``AD'' (anomaly detection), ``Gen'' (general), and ``Met'' (metrics).}
\vspace{2pt}
\footnotesize
\setlength{\tabcolsep}{3.2pt}
\renewcommand{\arraystretch}{1.02}
\begin{tabular}{llcccccc}
\toprule
\textbf{Task} & \textbf{Bench.} & \textbf{Turn} & \textbf{Multi} & \textbf{Agent} & \textbf{NL} & \textbf{Dom.} & \textbf{Eval} \\
\midrule
Fcst. & Monash \cite{godahewa2021monash}  & \xmark & \xmark & \xmark & \xmark & Gen. & Met. \\
Cls.  & UCR/UEA \cite{dau2019ucr,bagnall2018uea} & \xmark & \xmark & \xmark & \xmark & Gen. & Met. \\
AD    & NAB \cite{lavin2015nab}    & \xmark & \xmark & \xmark & \xmark & 3    & Met. \\
Fcst. & M4 \cite{makridakis2020m4}     & \xmark & \xmark & \xmark & \xmark & 6    & Met. \\
AD    & TSB-UAD \cite{paparrizos2022tsb} & \xmark & \xmark & \xmark & \xmark & Gen. & Met. \\
QA    & FinMTM \cite{zhang2026finmtm} & \cmark & \xmark & \cmark & \cmark & 1    & Jud. \\
\midrule
\textbf{All} & \textbf{TimeSage-MT} & \textbf{\cmark} & \textbf{\cmark} & \textbf{\cmark} & \textbf{\cmark} & \textbf{8} & \textbf{Mix} \\
\bottomrule
\end{tabular}
\label{tab:related_comparison}
\vspace{-10pt}
\end{wraptable}

Existing work relevant to our setting falls into four categories. First, LLM-based time series models such as LLMTIME \cite{gruver2023llmtime}, PromptCast \cite{xue2024promptcast}, Time-LLM \cite{jin2024timellm}, Chronos \cite{ansari2024chronos}, TimesFM \cite{das2024timesfm}, and Moirai \cite{woo2024moirai} show that language-model style architectures can capture temporal structure, while UniTS \cite{gao2024units} moves toward unified multi-task modeling. However, they are primarily \emph{model-centric}: they target one or several fixed tasks, rather than an interactive workflow that selects skills, executes multi-step analysis, and returns actionable decisions.

Second, AutoML systems for time series, including \texttt{auto.arima} \cite{hyndman2008forecast} and AutoGluon-TS \cite{shchur2023autogluon}, automate model selection and provide strong forecasting baselines. Yet they remain narrowly scoped to predefined predictive objectives and metric-based evaluation, without multi-turn clarification, open-ended natural-language requests, or cross-step reasoning over heterogeneous goals.

Third, general-purpose agent frameworks such as AutoGen \cite{wu2024autogen}, CAMEL \cite{li2023camel}, and TaskWeaver \cite{qiao2023taskweaver} demonstrate planning, tool use, and self-correction, but they are domain-agnostic. They do not provide a time-series-specific skill library, nor do they evaluate whether an agent can complete realistic multi-turn workflows. The comparison across benchmarks is summarized in table~\ref{tab:related_comparison}.

Finally, existing time series benchmarks focus on isolated canonical tasks: Monash \cite{godahewa2021monash} and M4 \cite{makridakis2020m4} for forecasting, UCR/UEA \cite{dau2019ucr,bagnall2018uea} for classification, and NAB \cite{lavin2015nab} and TSB-UAD \cite{paparrizos2022tsb} for anomaly detection. These benchmarks are valuable, but they are typically single-turn, single-task, and non-agentic. FinMTM \cite{zhang2026finmtm} introduces multi-turn and judge-based interaction, but it remains narrow in dialog depth, domain scope, and reasoning paradigms. 
In contrast, \textbf{TimeSage-MT} is designed to evaluate \emph{agentic time series analysis from conversation to decision}: multi-turn interaction, multi-task analytical trajectories, tool/skill orchestration, and domain-grounded outputs across 8 real-world domains and 4 complexity levels. This benchmark position follows directly from our broader benchmark and framework design goals.

\section{TimeSage-MT Benchmark}
TimeSage-MT is designed to evaluate time series analysis as a multi-turn analytical workflow rather than a single-step prediction task. Each task (see Figure \ref{Figure_3} and Appendix \ref{appendix_b}) includes the source time series, a visible data slice, a multi-turn dialogue, an analysis trajectory, reference code, numerical ground truth, reasoning annotations, and level-specific output requirements. These components make each interaction reproducible and verifiable. This enables us to assess whether an agent derives the correct answer, grounds analysis in executable evidence, and follows a valid reasoning path.

\subsection{Design Principles}
\label{section_3.1_design_principles}
The design of TimeSage-MT is driven by three requirements for evaluating time series agents in practical analytical workflows. \textbf{First}, tasks must be grounded in real-world time series data, since practical analysis depends on temporal properties such as trend, seasonality, missing values, outliers, regime shifts, uncertainty, and domain constraints. \textbf{Second}, the benchmark should reflect how analysis unfolds in practice:  users begin with exploration, request further analyses, synthesize evidence, and use accumulated results to support decisions. Thus, tasks are organized into 4 complexity levels, from L1 open exploration to L4 full decision paths, with each level representing a deeper analytical stage rather than a harder prompt. \textbf{Third}, multi-turn reasoning must be traceable and diagnosable. Thus, we propose a 9-capability taxonomy covering data inspection, method selection, parameter configuration, cross-turn memory, dependency tracking, adaptation, decision support, uncertainty calibration, and domain grounding. This design allows failures to be diagnosed beyond final-answer accuracy, revealing whether an agent breaks down because it misreads the data, chooses the wrong method, forgets prior evidence, mishandles uncertainty, or makes unsupported domain decisions. For a discussion of the rationale for the design and the capability taxonomy, please see the Appendix~\ref{appendix_a}.

\subsection{End-to-End Data Generation Pipeline}
\label{section_3.2_data_generation}
\begin{figure*}[!h]
\begin{center}
\includegraphics[width = \linewidth]{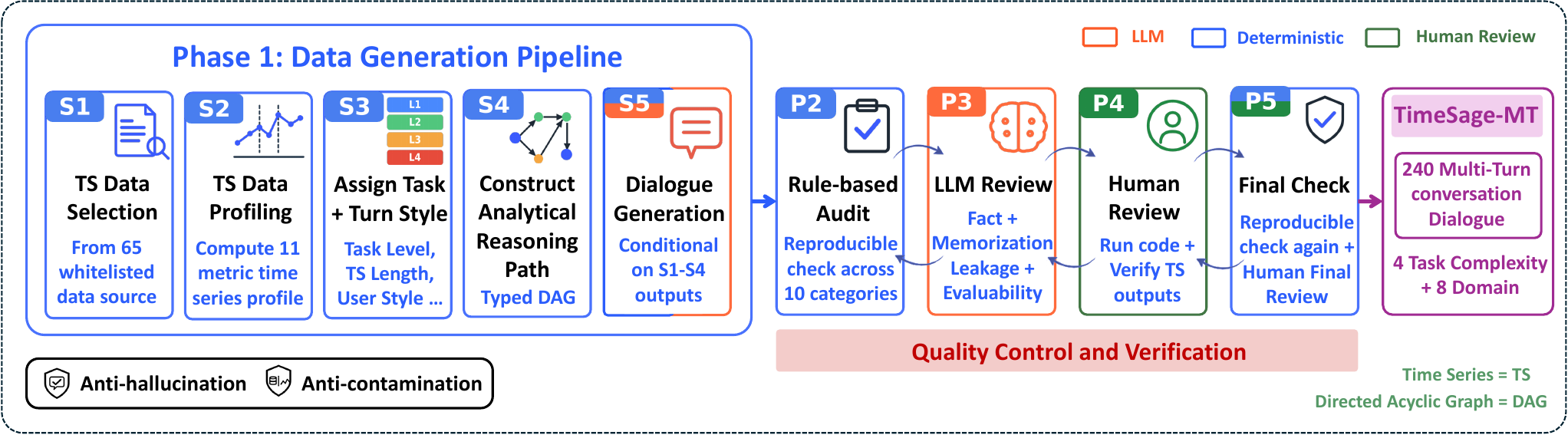}
\end{center}
\vspace{-2mm}
\caption{Reproducible construction and quality control pipeline for TimeSage-MT.}
\vspace{-2mm}
\label{Figure_2}
\end{figure*}

The left part of Figure \ref{Figure_2} illustrates the five-stage Phase 1 pipeline that turns raw time series into multi-turn analytical tasks (S1–S4 deterministic, S5 LLM-powered). \textbf{S1 (Data Selection)} curates time series from 65 whitelisted real-world sources, validating license, format, and level-appropriate length. Sources appearing in the pretraining corpora of the time series foundation model are flagged for contamination-aware construction and evaluation. \textbf{S2 (Profiling)} applies an 80/20 visible/held-out split, using only the visible window for construction and reserving the rest for evaluation. Each source receives an 11-metric profile that covers frequency, dimensionality, stationarity, seasonality, missingness, temporal complexity, and feasible task families. \textbf{S3 (Task Assignment)} uses profile-derived eligibility constraints to assign levels, dialogue lengths, user styles, domains, and analytical families, while enforcing a quota of 60 tasks per level. \textbf{S4 (Reasoning Path)} turns each profiled source-task assignment from S1--S3 into a reasoning graph specifying ordered analyses, dependencies, and expected outputs: L2 uses linear chains, L3 combines multiple domain-grounded analyses into an integrated summary, and L4 extends these with decision drivers, uncertainty calibration, and a final structured decision output. \textbf{S5 (Dialogue Generation)} prompts Claude Opus 4.6 \cite{anthropic2026claudeopus46} with the profile, assignment, reasoning path, visible data preview, and skill registry to produce multi-turn dialogues. Verifiable answer targets are computed by deterministic extractors running reference code over the underlying time series, not taken from the dialogue, and neither the code nor the targets are exposed to evaluated agents. Please refer to Appendix~\ref{appendix_b} for more details.

\subsection{Quality Control and Verification}
The right part of Figure \ref{Figure_2} shows the 4-stage quality control pipeline (Phasses 2-4) through which each generated task is passed to reduce hallucinations, leakage, and unsupported targets. The \textbf{reproducibility audit} (P2) applies 46 deterministic checks in 10 categories, including data integrity, split integrity, assignment validity, graph validity, reference-code execution, gold-target provenance, narrative grounding, metadata consistency, cross-task consistency, and knowledge-pack integrity. The \textbf{LLM review} (P3) flags inconsistencies in dialogue, analysis trajectory, and answer targets, and screens for residual memorization or open-loop evaluability issues. Human \textbf{ review} (P4) is conducted through an annotation dashboard that displays the source data, dialogue, reasoning path, reference code, gold fields, and review status in a unified interface, allowing reviewers to inspect and refine each task. The \textbf{final check} (P5) reruns audits, re-executes reference code in a sandbox, and routes any failure back for regeneration. The annotation dashboard is released as a reusable component, enabling future benchmark extensions to add new domains, task types, and conversation styles under the same review process. The complete checks and dashboard details are provided in the Appendix~\ref{appendix_b}.

\vspace{-2mm}
\begin{figure*}[!h]
\begin{center}
\includegraphics[width = \linewidth]{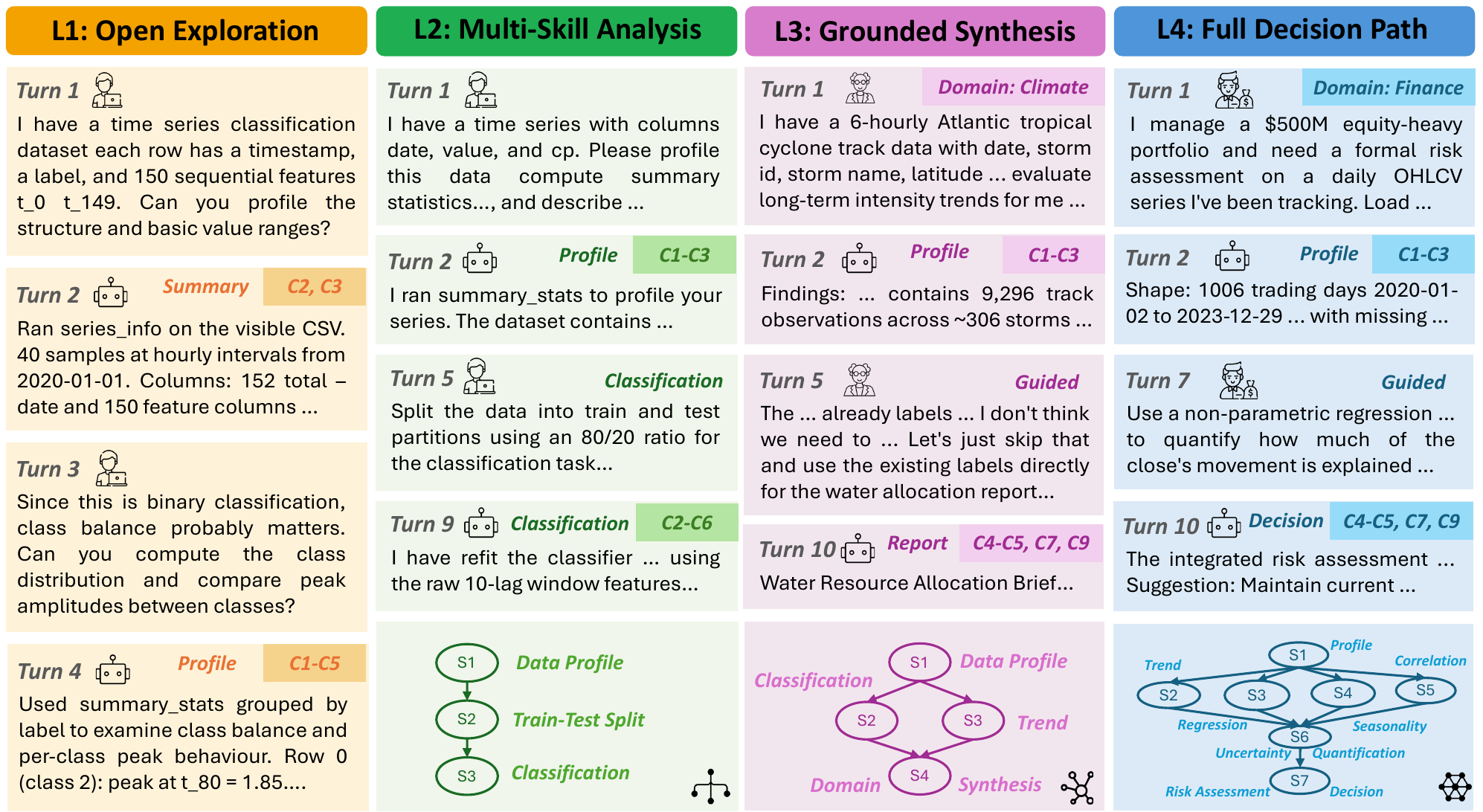}
\end{center}
\vspace{-3mm}
\caption{Representative multi-turn conversations from each of the
4 difficulty levels.}
\vspace{-2mm}
\label{Figure_3}
\end{figure*}

\subsection{Dataset Statistics}
\begin{wrapfigure}{r}{0.4\linewidth}
\vspace{-8pt}
\centering
\includegraphics[width=0.8\linewidth]{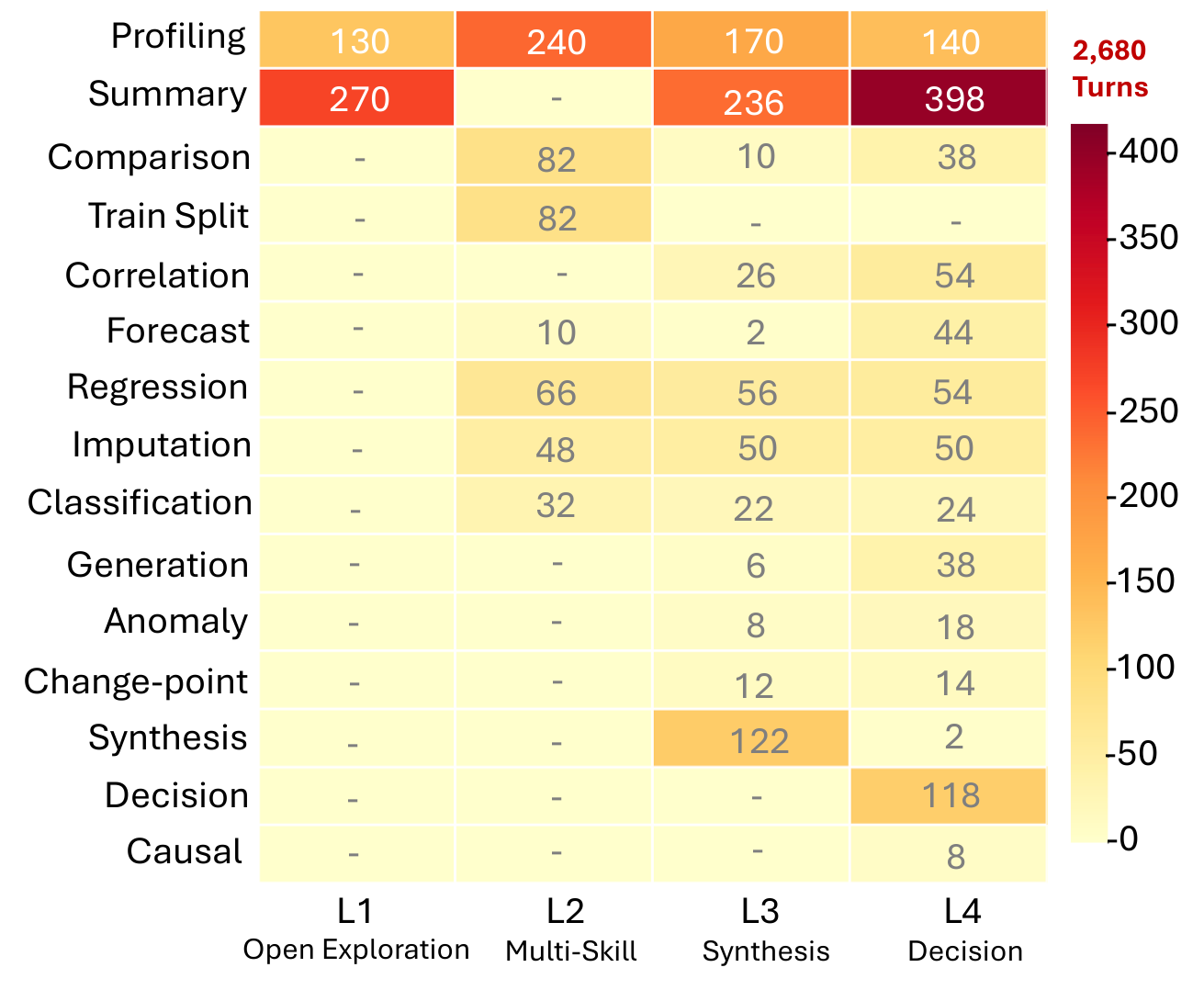}
\caption{Distribution of time series analytical task families across L1-L4 levels.}
\label{Figure_4}
\vspace{-8pt}
\end{wrapfigure}
TimeSage-MT comprises 240 tasks and 2,680 dialogue turns, evenly distributed across 4 difficulty levels (60 tasks each): L1 open exploration, L2 multi-skill analysis, L3 grounded synthesis, and L4 full decision path. Representative conversations and their reasoning graphs are illustrated in Figure \ref{Figure_3}. L1–L2 contain 120 general-purpose tasks; L3–L4 contain 120 domain-grounded tasks that span eight domains: healthcare (18), energy (17), manufacturing (16), climate (16), finance (15), telecommunications (14), transportation (13), and retail (11). Tasks cover a wide range of temporal resolutions, including 100 \, Hz, per-sample, per-connection, 1-minute, 5-minute, 15-minute, hourly, daily, monthly and irregular event-based streams. The dialogue style is partitioned into 120 open tasks, 80 guided-correct tasks, and 40 guided-wrong tasks. Conversation length ranges from 4 to 20 turns, and reasoning-graph depth scales from 3 nodes at L2 to 10 nodes at L4. Figure~\ref{Figure_4} shows that the benchmark's coverage of time series analysis task families broadens across levels, from descriptive turns to varied analytical and decision-oriented workflows.

\section{Benchmarking Agentic Time Series Systems}
\subsection{Evaluation Protocol}
Our evaluation protocol is designed around the following commitments: \ding{172} scoring is deterministic for every dimension with a verifiable ground truth, \ding{173} every leaderboard score is reproducible from a structured trace we release for each run, and \ding{174} all grading is reported at the per turn level so failures and successes can be located within the dialogue.

\paragraph{Outcome scoring.}
The Outcome track grades each task along 5 dimensions from numerical correctness to decision soundness. Three of the five are graded by deterministic rules. \textbf{Numerical accuracy} compares the agent's stated values against gold targets that we extract at task construction by running reference code in the sandbox. \textbf{Factual verification} matches the agent's stated profile attributes against gold values across both the structured output and the dialogue narrative. \textbf{Code correctness} runs the agent's Python code in a sandbox and compares its output with the reference. The remaining dimensions, \textbf{analytical quality} and \textbf{decision quality}, are scored by an LLM judge over the dialogue narrative and the structured decision payload required on level 4 tasks. The total per task is the mean of the applicable tracks (4 for L1 to L3, 5 for L4); inapplicable tracks are excluded from the mean. The evaluation criteria for each dimension are detailed in the appendix \ref{app:outcome_evaluation}.

\paragraph{Capability scoring.}
The Capability track assesses agent behavior across the analytical workflow. We map each agent's reasoning trace onto the 9 capabilities defined in Section \ref{section_1_introduction} and grade them through 28 sub-checks, of which 26 use deterministic rules and only 2 (C7 decision making and C9 domain reasoning) require an LLM judge. Each sub-check is scored per task and per turn, with applicability set by task level: C4 Remember and C5 Chain activate at L2, where cross turn dependencies are introduced, and C7 Decide, C8 Calibrate, and C9 Ground activate at L3 and L4 once decisions enter the dialogue. Capability scores are diagnostic, not a leaderboard claim: a high C5 paired with a low C1 tells the researcher that the agent threaded its methods correctly but misread the underlying data, which is a distinction the composite total cannot show. The complete applicability matrix and sub-check definitions appear in the appendix \ref{app:capability_evaluation}.

\paragraph{LLM judge protocol.} Analytical quality, decision quality, C7 decision making, and C9 domain reasoning are graded by an LLM judge \footnote{All reference targets are produced by executable code rather than an LLM.}. We use GPT-5.4 \cite{openai2026gpt54} as the judge in all agents and system settings. All judge scores, prompts, and rubrics are released alongside the leaderboard so future work can apply alternative judge configurations. The judgment prompt is presented in the appendix \ref{app:capability_LLM_judge}.

\begin{figure*}[!h]
\begin{center}
\includegraphics[width = \linewidth]{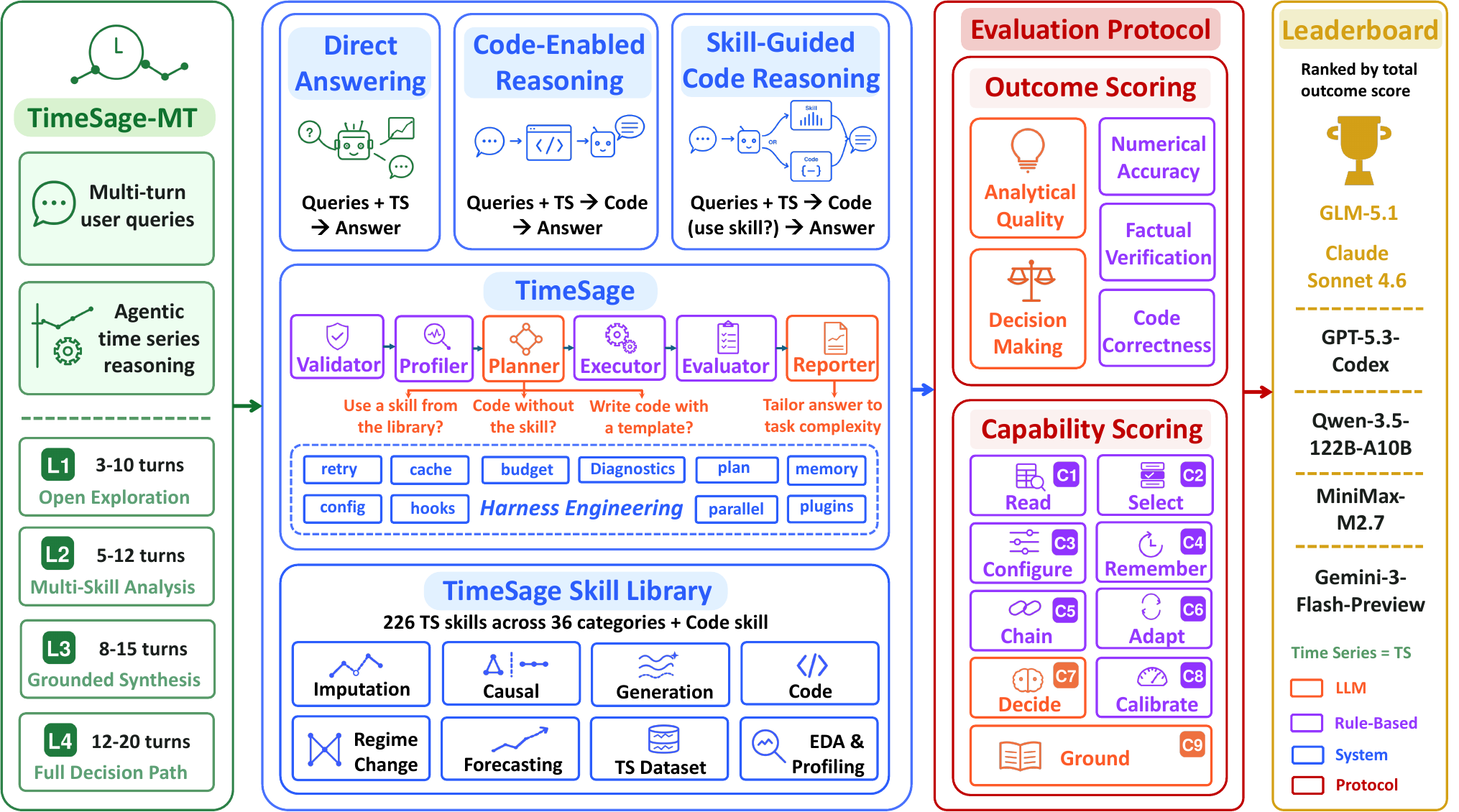}
\end{center}
\vspace{-3mm}
\caption{Overview of agentic system settings and evaluation protocol.}
\vspace{-2mm}
\label{Figure_5}
\end{figure*}

\subsection{Implementation Details}
\paragraph{System settings.}
The settings below progressively layer structure onto a frontier model, from no execution environment (Direct Answering) to a structured agentic system (TimeSage). We use this ladder to characterize where added structure helps and where it does not. Every run produces a structured trace from which the leaderboard scores can be reproduced. The complete system architecture and prompts are provided in Appendix~\ref{appendix_d}.

\begin{itemize}[leftmargin=*]
    \item \textbf{Direct Answering.} The model receives the user turn and time series data, with no computational tools. Responses are produced from the model's prior knowledge and direct inspection of the data.
    \item \textbf{Code-Enabled Reasoning.} The model operates within an execution environment that allows it to access the time series, compute statistics, and perform analyses end to end through generated code.
    \item \textbf{Skill-Guided Code Reasoning.} The model has access to the TimeSage skill library, where each skill includes a code template with descriptions, and can select a skill or generate its own code.
    \item \textbf{TimeSage.} A fully orchestrated agentic pipeline of 6 stages, of which 2 use an LLM (\textit{Planner} and \textit{Reporter}) and 4 are deterministic (\textit{Validator}, \textit{Profiler}, \textit{Executor}, and \textit{Evaluator}). The Planner decomposes each user turn into a workflow over the TimeSage skill library, which contains 226 skills across 36 categories and a general purpose code skill for cases outside the set. The Executor runs the workflow under deterministic quality gates and the Reporter adapts the response to the task. Figure \ref{Figure_5} shows the architecture, and Appendix \ref{appendix_e} gives details per stage and skill library.
\end{itemize}

\paragraph{Models.}
We evaluate 6 backbones drawn from closed and open source families: Claude Sonnet 4.6 \cite{anthropic2026sonnet46}, GPT-5.3-Codex \cite{openai2026gpt53codex}, GLM-5.1 \cite{zai2026glm51}, Qwen-3.5-122B-A10B \cite{qwen2026qwen35122ba10b}, MiniMax-M2.7 \cite{minimax2026m27}, and Gemini-3-Flash-Preview \cite{google2026gemini3flashpreview}, all run at temperature 0 with reasoning effort set to high where supported (medium for GPT-5.3-Codex). Code-Enabled Reasoning anchors the leaderboard on the full 240 task corpus, while the other settings run on a 100 task paired subset stratified across tiers and domains using Qwen-3.5-122B-A10B and GPT-5.3-Codex, so every system level delta reported in Section~\ref{main_leaderboard_performance} and \ref{section_4.4_key_observations} is over identical task IDs and judge assignments.

\subsection{Leaderboard Results}
\label{main_leaderboard_performance}

\begin{figure*}[!h]
\begin{center}
\includegraphics[width =0.9\linewidth]{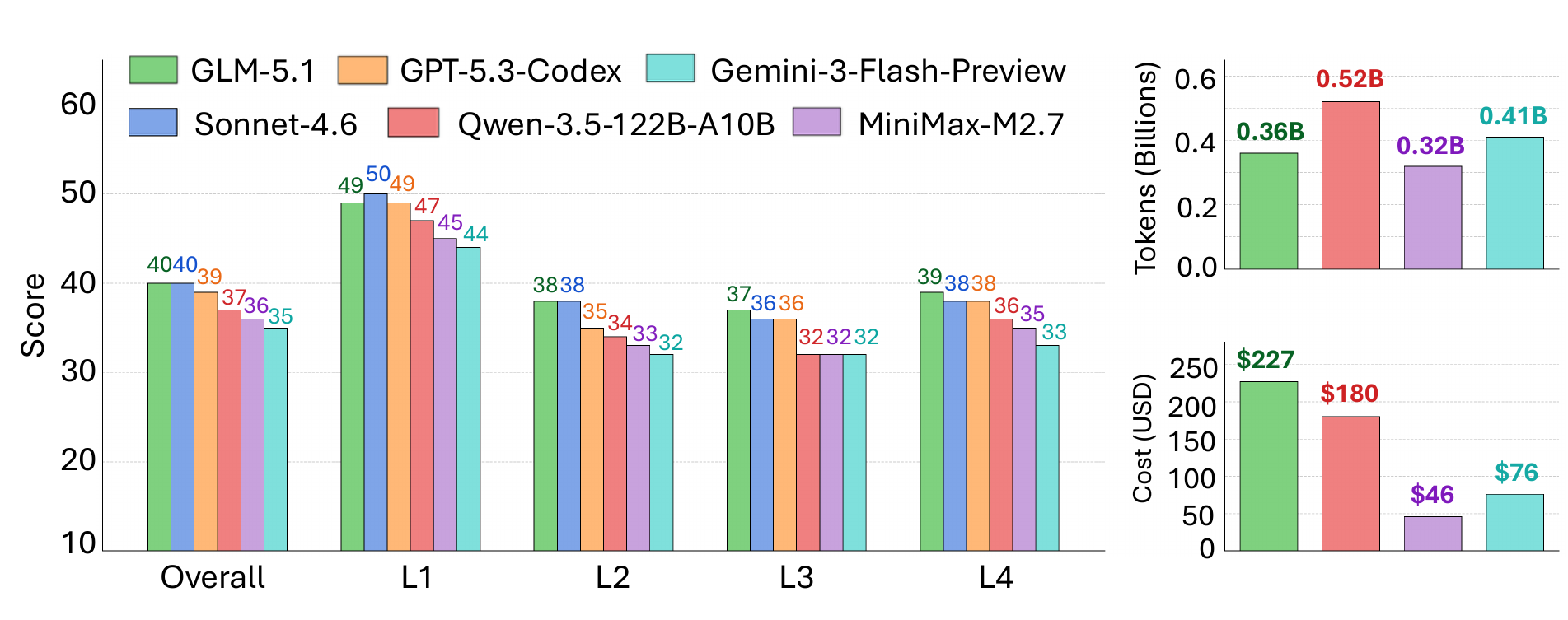}
\end{center}
\vspace{-3mm}
\caption{\textbf{Left}: Outcome scores across difficulty tiers L1–L4 and overall. \textbf{Right}: Token cost.}
\vspace{-2mm}
\label{Figure_6}
\end{figure*}

We evaluate six frontier LLMs on the 240-task corpus under Code-Enabled Reasoning to establish the main leaderboard.
Figure~\ref{Figure_6} presents the tier-wise and overall results, highlighting three key findings:
\textbf{\ding{172} Performance drops with dialog depth.} All models perform best on L1, but drop by 10--15 points at L2, with further slight declines at L3 and L4, showing that TimeSage-MT distinguishes shallow exploration from action-oriented, multi-turn reasoning.
\textbf{\ding{173} Frontier LLMs are closely matched.} The six models fall within a narrow 35--40 overall range, with GLM-5.1 and Sonnet-4.6 leading at 40 and GPT-5.3-Codex at 39. 
The absence of a clear winner suggests that performance degradation as dialog depth increases is shared across model families and may require better harness design, not only stronger backbones.
\textbf{\ding{174} Higher cost does not imply better quality.}\footnote{GLM-5.1, MiniMax-M2.7, Qwen-3.5-122B-A10B, and Gemini-3-Flash-Preview are accessed through OpenRouter. Sonnet-4.6 and GPT-5.3-Codex are accessed through CLI tools, which do not expose per-call token usage; their cost and token counts are omitted.}
Among the four models with reported costs, spending varies by nearly 5$\times$, while scores differ by only a few points. This suggests that the inference budget is not the bottleneck.

\begin{figure}[!t]
    \centering
    \begin{subfigure}[t]{0.38\linewidth}
        \centering
        \includegraphics[width=\linewidth]{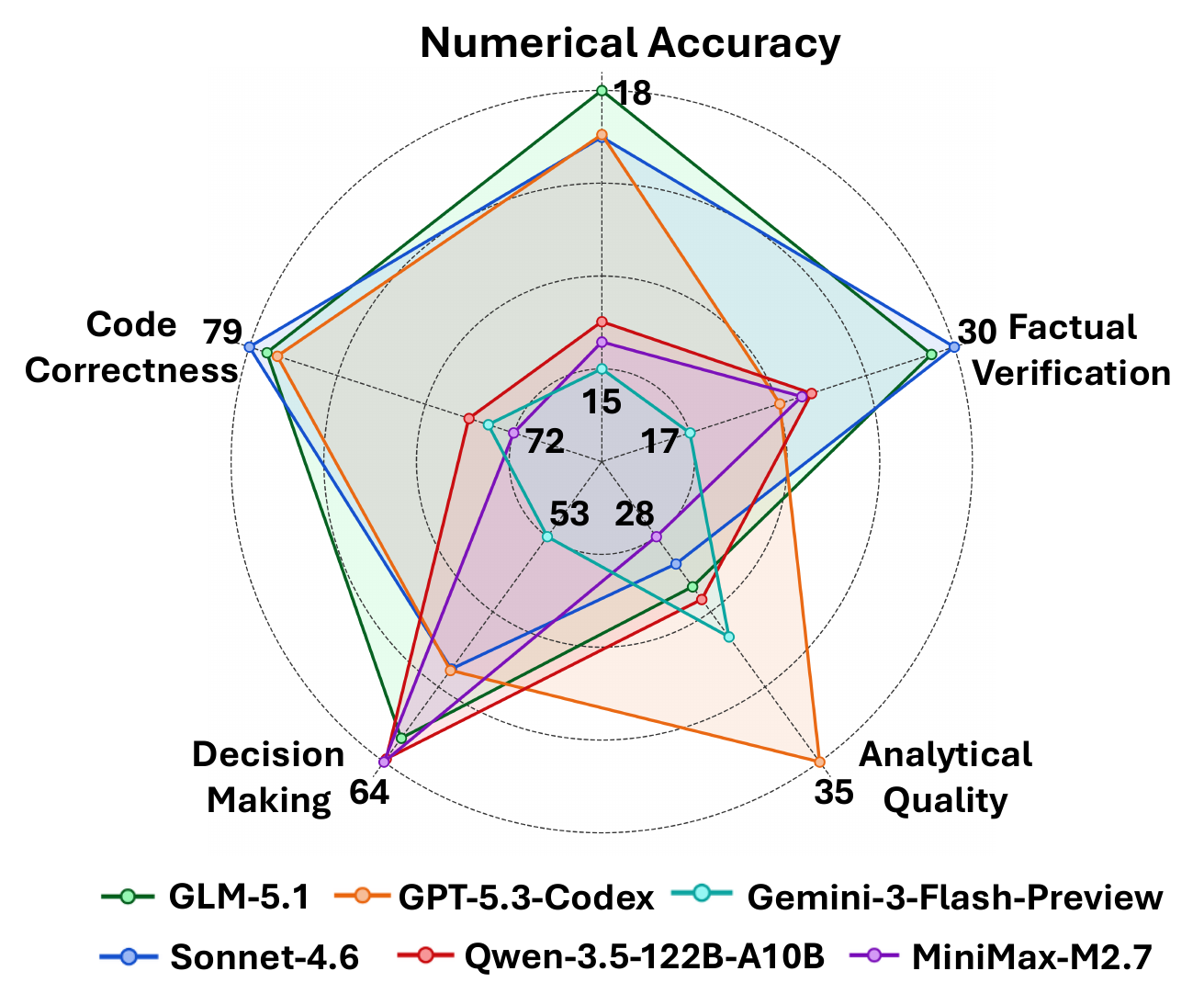}
        \caption{Outcome scores across five dimensions}
        \label{Figure_7:radar}
    \end{subfigure}
    \hfill
    \begin{subfigure}[t]{0.58\linewidth}
        \centering
        \includegraphics[width=\linewidth]{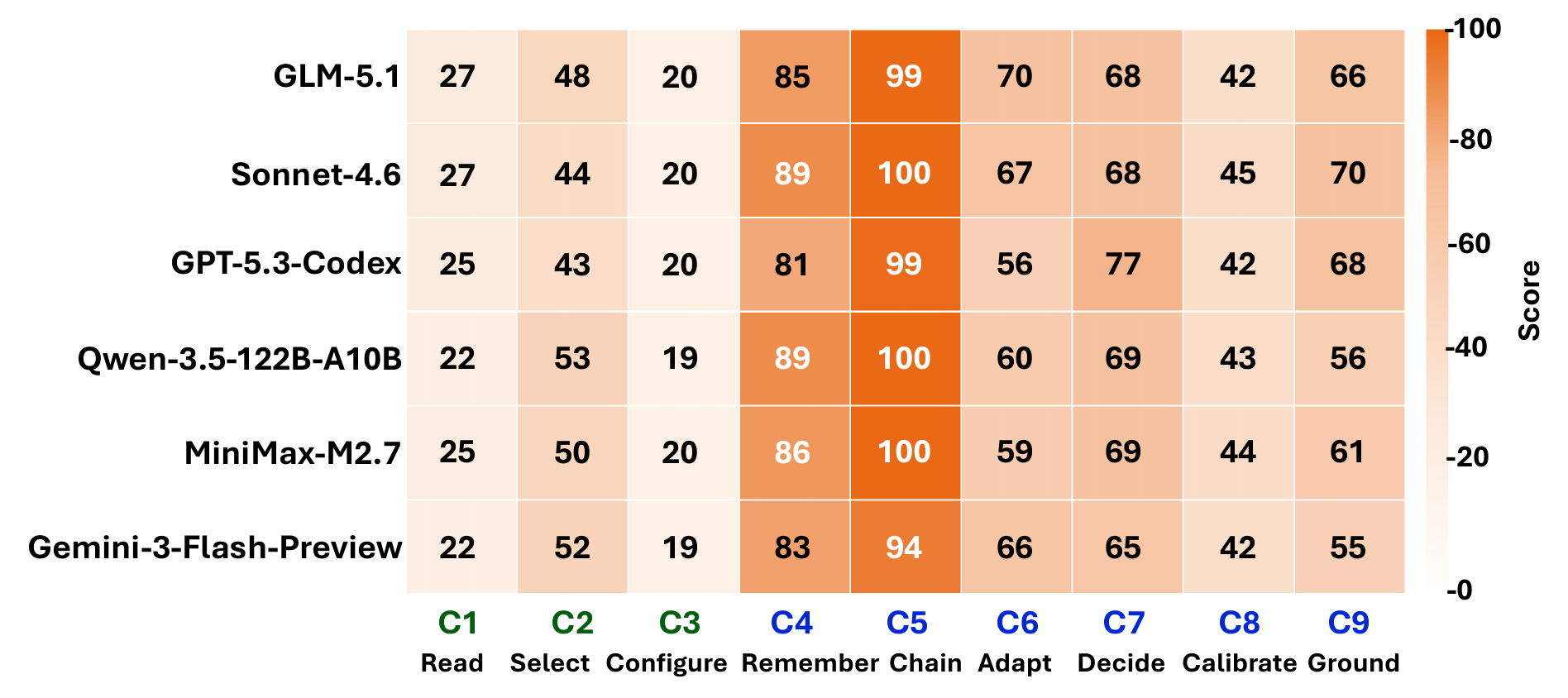}
        \caption{Outcome scores across nine capabilities}
        \label{Figure_7:heatmap}
    \end{subfigure}
    \caption{Diagnostic decomposition of LLM time series reasoning across five outcome dimensions (a) and nine fine-grained capabilities (b).}
    \label{Figure_7}
\end{figure}

Figure~\ref{Figure_7:radar} reveals a clear gap between execution and reliable reasoning: 
\textbf{\ding{172} Code execution is not the bottleneck.} 
Most models achieve high Code Correctness but score much lower on other dimensions, indicating that runnable code alone does not ensure trustworthy, target-grounded analysis. 
\textbf{\ding{173} Models fail along different dimensions,} GLM-5.1 and Sonnet-4.6 are more balanced, GPT-5.3-Codex is stronger in code and analytical writing but weaker in factual verification, while Qwen-3.5-122B and MiniMax-M2.7 perform better in decisions than in numerical grounding.
Figure~\ref{Figure_7:heatmap} further localizes these failures:
\textbf{\ding{174} Memory and chaining are relatively strong.} High C4 Remember and C5 Chain scores show that models can usually retain prior results and follow dependencies.
\textbf{\ding{175} Analytical grounding remains weak.} Low C1 Read and C3 Configure scores reveal persistent errors in reading time series properties and setting appropriate analysis parameters.


\subsection{Key Observations}
\label{section_4.4_key_observations}

In this section, we move from benchmark-scale rankings to controlled system-level analyses. To eliminate sampling confounds, all comparisons use a 100-task paired subset stratified across the four tiers and eight domains, with task identifiers and judge assignments held fixed across configurations.

\input{table/observation_1_2}

\subsubsection{Observation \#1: Code-Enabled Reasoning Improves Time Series Analysis}

We examine whether code execution improves time series reasoning over direct natural-language answering.
Table~\ref{tab:ob_1}a shows that Code-Enabled Reasoning consistently outperforms Direct Answering across all dimensions and difficulty tiers. The largest gap is in Code Correctness, where Direct Answering scores 0 because it does not produce executable evidence. Without execution-based grounding, even frontier models often fabricate numbers or unsupported statistics, and these errors propagate to the final analysis and recommendations. This is especially harmful in time series analysis, where small numerical errors can alter trends, hide anomalies, or reverse decisions.

\subsubsection{Observation \#2: Skill Library Provides Additional Gains}

We compare Code-Enabled Reasoning with Skill-Guided Code Reasoning to test whether predefined analytical routines improve standard code execution.
Table~\ref{tab:ob_1}b shows model-dependent gains. GPT-5.3-Codex improves from 39 to 45 overall, through higher Analytical Quality (35 to 61) and Decision Making (63 to 76), suggesting that skills help organize more coherent analyzes. In contrast, Qwen-3.5-122B improves marginally from 37 to 38, with most dimensions unchanged. A skill library is helpful when the model can invoke the right routines and integrate their outputs effectively.

\subsubsection{Observation \#3: Structured Pipeline Improves Grounding but Restricts Expressiveness}

\begin{figure*}[!t]
\begin{center}
\includegraphics[width =\linewidth]{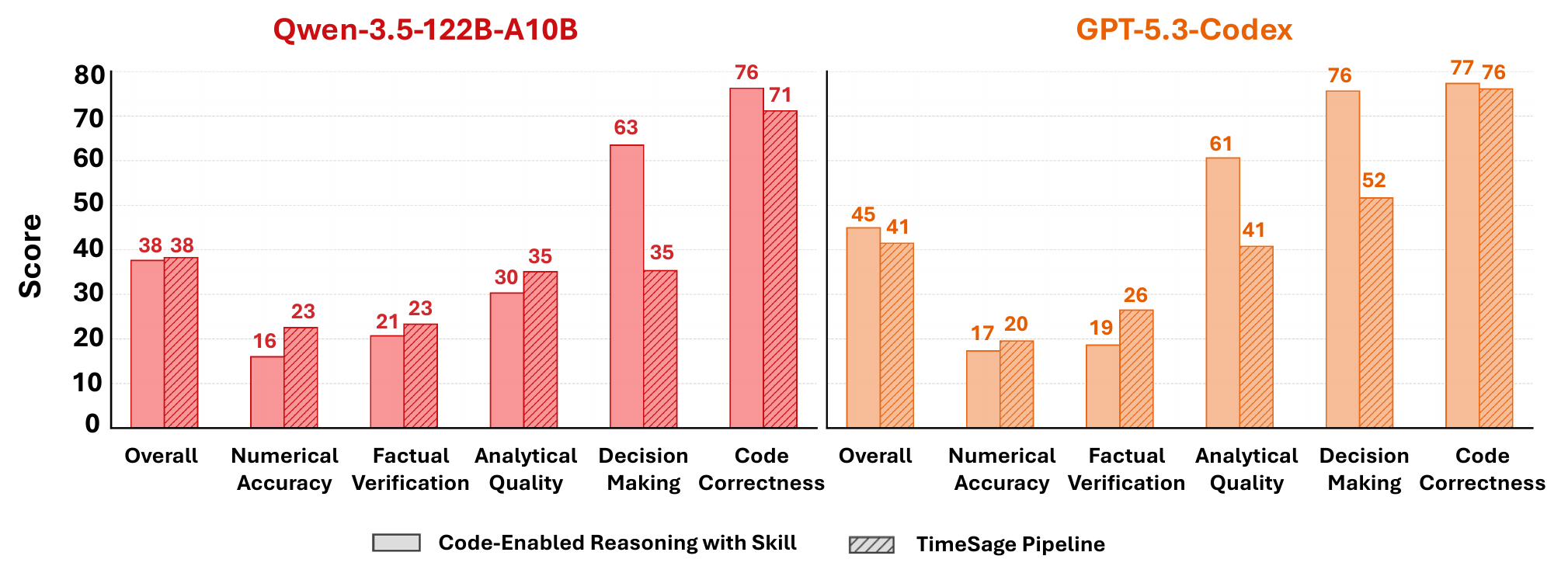}
\end{center}
\vspace{-3mm}
\caption{
Effect of the agentic pipeline  over Skill-Guided Code Reasoning on Qwen-3.5-122B-A10B (left) and GPT-5.3-Codex (right) across the five outcome dimensions and overall.
}
\vspace{-6mm}
\label{Figure_8}
\end{figure*}

Finally, we compare skill access with structured pipeline to test whether explicit planning and validation further improve reasoning.
Figure~\ref{Figure_8} shows a clear trade-off. For Qwen-3.5-122B, TimeSage improves Numerical Accuracy, Factual Verification, and Analytical Quality, but reduces Decision Making and Code Correctness, yielding only a small overall gain. This suggests that orchestration helps weaker backbones stay grounded, but may limit response flexibility.
For GPT-5.3-Codex, TimeSage lowers all dimensions, especially Analytical Quality and Decision Making. Since GPT-5.3-Codex already runs in an agentic CLI harness, an additional workflow layer may introduce planning conflicts. These results show that a structured agentic pipeline is not a universal upgrade: it benefits backbones that need scaffolding to stay grounded but can dampen models whose own agentic harness is already strong, motivating the harness-vs-LLM trade-offs.




\section{Conclusion and Future Works}

We introduced TimeSage-MT, a multi-turn benchmark for evaluating agentic time series reasoning. TimeSage-MT contains 240 tasks and 2,680 dialogue turns across 8 domains and 4 complexity levels, with verifiable targets, reference code, reasoning trajectories, and capability-level annotations. Our evaluation shows that code execution and skill libraries improve performance, but frontier LLM agents still struggle with two core weaknesses: numerical accuracy and analytical grounding in time series analysis. The agentic pipeline introduces a clear trade-off: it can improve numerical accuracy and grounding but may hurt analytical quality and decision making when its structured workflow constrains the model's native reasoning behavior. These findings suggest that reliable time series agents require not only stronger LLMs but also better harness design rather than simply adding more human-inspired pipeline stages. Future work should develop adaptive orchestration mechanisms that preserve model flexibility while improving numerical verification, uncertainty calibration, and domain-grounded decision making. We hope that TimeSage-MT will support future work on trustworthy and auditable agents for real-world time series analysis.



\newpage

{
\small
\bibliographystyle{unsrt}
\bibliography{reference}
}

\newpage

\appendix

\section{Design Rationale for TimeSage-MT Benchmark}
\label{appendix_a}
Existing time series benchmarks ask one question per task: forecast the next 24 steps; generate a time series; or find anomalies in a stream. However, real analysis is a workflow. An analyst inspects the data; picks methods that match its properties; runs dependent analyses; revises when the user redirects; and writes up a recommendation that takes uncertainty seriously. \textit{TimeSage-MT} is designed to evaluate that \textbf{full workflow} rather than its end-points. In order to do this, we decompose agentic time series reasoning along two orthogonal axes: \textbf{task complexity} (how deep the analytical chain runs, L1-L4) and \textbf{capability} (what kind of skill the agent must demonstrate, C1-C9). Every task is paired with a structured gold trajectory so each axis can be scored independently. A low composite score can be diagnosed (was it a Read failure or a Calibrate failure?) rather than left as a black box.

\subsection{Capability Taxonomy (C1--C9)}
The benchmark uses 9 capabilities, denoted C1-C9.  A capability is not a model feature. It is an observable behavior that can be scored from the agent's turns, tool traces, code, and final output.  Table~\ref{tab:capability-taxonomy} summarizes the operational definition of each capability.

\input{table/appendix_table/appendix_a/1_capability}

\subsection{Task-Level Taxonomy (L1-L4)}
The level is the principal complexity axis. L1 is open-ended exploratory questions at 3-10 turns; L2 layers on multi-step chaining (5-12 turns); L3 returns a written synthesis grounded in 8 high-stakes domains (8-15 turns); L4 returns a structured decision with calibrated confidence at the end of a longer trajectory (12-20 turns). Each level holds 60 tasks ($60 \times 4 = 240$), partitioned evenly into short / medium / long length sub-categories of 20 tasks each. The length varies with complexity but does not dominate it. Table \ref{tab:task-levels} illustrates the task-level design. 

The task levels are crossed with three analyst styles. In the open style, the user asks natural analytical questions without telegraphing the answer. In the guided-correct style, the user gives helpful hints that point to a sound method. In the guided-wrong style, at least one user turn contains a plausible but misleading suggestion, and the gold agent must push back using evidence. The style distribution is balanced within each level: 30 open tasks, 20 guided-correct and 10 guided-wrong tasks per level.

\input{table/appendix_table/appendix_a/2_task}

\section{TimeSage-MT Benchmark Construction Pipeline}
\label{appendix_b}
The benchmark is built using a five-phase pipeline (Figure~\ref{Figure_2} of Section \ref{section_3.2_data_generation}). Phase~1 is the generation phase, where it creates candidate tasks through five sub-stages S1-S5. Phases 2--5 are quality control. P2 performs deterministic reproducibility audits.  P3 performs cross-family LLM review.  P4 performs assigned human review in an annotation dashboard.  P5 finalizes the corpus by combining human approval state with deterministic reproducibility checks and, where configured, LLM review and reference-code execution. Table~\ref{tab:app_pipeline_overview} summarizes which phases use an LLM, which are deterministic, and where human review enters.

\input{table/appendix_table/appendix_b/pipeline}

\subsection{P1.S1: Data Selection}

S1 is deterministic. It builds a whitelisted source manifest from real-world time series datasets. The construction scripts validate that a source can be retrieved or loaded locally, has a permitted license, parses as a CSV-like table, includes a time axis, includes at least one numeric signal, and is long enough for the target levels. The published benchmark is generated from 65 approved source collections. The source-whitelist acceptance checklist shown in Table \ref{tab:whitelist_check} is applied before a candidate source can enter the benchmark pool. 

Contamination risk is recorded as metadata. TimeSage-MT does not train a time series foundation model on these sources.  Instead, contamination risk means that a source is documented or strongly likely to overlap with the pretraining corpus of at least one external time series foundation model. S1 records the flag and the corresponding contamination sources; S3 propagates this metadata to task assignments; P2 and P5 verify that the fields are present. This enables the evaluation to report clean-only, contamination-risk-only, and stratified scores. In the published benchmark, 15 of the 65 approved source collections and 65 of the 240 final tasks carry contamination risk.

The complete whitelist of data sources, including URLs, domain categories, and contamination-risk metadata, is available at \url{https://github.com/TimeSage-Series/TimeSage-MT}.

\input{table/appendix_table/appendix_b/P1/1_whitelist_checklist}

\subsection{P1.S2: Time Series Profiling}

S2 is deterministic. For each source, it creates an 80/20 visible/held-out split. The visible CSV contains only the first 80\% visible slice; the final 20\% held-out slice is reserved for gold extraction and final scoring.  S2 also computes the time series profile used by later stages: inferred frequency, length, dimensionality, stationarity statistics, trend and seasonality strength, missingness, outlier rate, lag-1 autocorrelation, noise-to-signal ratio, and feasible task families. The profile checklist below (Table \ref{tab:profile_checklist}) summarizes the deterministic checks used to construct the visibility contract and the source-level time series profile.

\input{table/appendix_table/appendix_b/P1/2_profile_checklist}

\subsection{P1.S3: Task Assignment}

S3 is deterministic. It samples level, length bucket, user style, domain, source, and task family under fixed quota. Each level contains 60 tasks: 20 short, 20 medium, and 20 long.  L1 and L2 are general analytical tasks; L3 and L4 include domain-grounding constraints so that synthesis and decision-making can be evaluated. Table \ref{tab:task_assign_checklist} summarizes the assignment checklist.

\input{table/appendix_table/appendix_b/P1/3_task_assign}

\subsection{P1.S4: Reasoning Path Construction}

S4 is deterministic. It converts each assignment into a typed analytical DAG.  Formally, each S4 graph is $G=(V,E,\tau_V,\tau_E,\alpha)$, where $V$ is a finite set of analytical nodes, $E\subseteq V\times V$ is a directed acyclic dependency relation, $\tau_V$ assigns each node a node type, $\tau_E$ assigns each edge a dependency type, and $\alpha$ stores node and edge attributes used by generation and evaluation. Node types include profile, transform, forecast, anomaly, classification, regression, imputation, change-point, causal, counterfactual, uncertainty, synthesis, and decision. Edge types include data-preparation, parameter dependence, evidence dependence, comparison, correction, synthesis, and decision-support. The typed DAG is therefore not only an ordering constraint: it specifies which analytical operation is expected, which earlier outputs it may depend on, and how the dependency should be interpreted by dialogue generation and later capability scoring. Table \ref{tab:reasoning_path_checklist} summarizes the checklist used to build reasoning paths.

The required reasoning-path shape varies by level.  L1 has no required DAG. L2 receives a short chain of 2--4 steps. L3 receives a 4--6 step domain-grounded graph with a synthesis node. L4 receives a 6--10 step graph ending in a decision node. Each node is tagged with required skills, allowed alternatives, capability tags, expected output type, and edge rationales. The examples below illustrate the L2 analytical-chain case and the L4 decision-template case.

For L2, the graph is a deterministic analytical chain rather than a domain decision template. For example, an L2 task may ask the agent to inspect a series of visible energy-loads, identify missingness and outliers, resample or clean the series, and then compute a short diagnostic summary. The resulting
path might be \texttt{profile} $\rightarrow$ \texttt{preprocess} $\rightarrow$ \texttt{anomaly} $\rightarrow$ \texttt{summary}. Each downstream node depends on the previous node's output: the preprocessing step uses the profile result to choose a frequency and missing-value treatment, the anomaly step operates on the cleaned series, and the final summary cites the cleaned statistics and anomaly evidence. This level therefore tests whether the agent can chain analytical operations across turns without requiring domain synthesis or a final decision.

For L3 and L4, the pipeline first assigns each task to a domain scenario template considering of the task identifier and domain. These templates specify a domain framing, a title template, a decision or
investigation type, and a set of sub-question categories. For L3, the template is used to construct a domain-grounded investigation graph that ends in a synthesis node. For L4, the same template family is passed through the decision-making framework: the template's decision type determines the primary
decision driver, such as forecasting, risk assessment, anomaly triage, root-cause analysis, policy evaluation, resource allocation, buy/sell/hold, or trend interpretation. S4 then instantiates analytical nodes from the skill-aware node catalog, filters nodes whose output type cannot be scored on the selected source, adds uncertainty and terminal decision nodes for L4, wires typed edges, normalizes node weights, and runs the structural and capability validators before saving the graph.

As a concrete example, an L4 finance task may be assigned to a \texttt{buy\_sell\_hold} decision template. The resulting reasoning path begins with a profile node that reads the visible price, return, and volatility structure; continues through analytical nodes for trend or regime characterization, risk estimation, and forecast or scenario evidence; and then adds an uncertainty node that calibrates the decision driver. The final answer is therefore not only a narrative recommendation: it must provide a structured decision set such as \{\texttt{buy}, \texttt{hold}, \texttt{sell}\}, a selected action, a magnitude or timing field when applicable, a confidence range, and rationale keys that cite upstream numerical evidence. Other L4 templates use the same typed-graph pattern but change the driver: anomaly-triage tasks end in an escalation or intervention decision, policy-evaluation tasks end in an intervention-effect decision, and resource-allocation tasks end in an allocation or scheduling decision. The details of this template can be accessed in the codebase.

\input{table/appendix_table/appendix_b/P1/4_reasoning_graph}

\subsection{P1.S5: Dialogue Generation and Gold Extraction}

In S5, the dialogue narrative is generated by an LLM, while visibility artifacts, gold targets, correction-turn metadata, and quality gates are deterministic. We prompt Claude Opus~4.6 (reasoning effort is set to high) with the S2 profile, the S4 reasoning graph, the data preview, and the canonical skill registry, and request the output containing the multi-turn dialogue grounded in the actual data. Every agent turn must declare expected skills, expected parameters, reference code (5--25 lines of Python that runs against the visible time series), and a verification spec (key finding and finding verify). Verifiable answer targets are produced by deterministic extractors that run the reference code on the held-out series, never by the LLM. The dialogue text and the gold target therefore come from independent sources, removing a large class of hallucination risk. Table \ref{tab:dialogue_generation_checklist} shows the checklist for S5.

\paragraph{S5 LLM prompt excerpt.}
For readability, we report the core dialogue-generation instruction rather than the complete template. The released prompt additionally specifies the output schema, validation constraints, recovery behavior for malformed outputs, and failure cases handled by the deterministic post-processor. The full prompt is released at \url{https://github.com/TimeSage-Series/TimeSage-MT}. The following excerpt gives the central instruction used by the LLM in S5:

{\small
\begin{verbatim}
System:
You are an expert conversation designer specializing in time series analytics.
You write realistic, grounded multi-turn dialogues between a human analyst and
an AI agent called Time-Sage. The agent always cites specific numbers from the
data and names the analytical skills it uses. Output ONLY valid JSON.

User:
Generate a {target_turn_count}-turn dialogue for a Time-Sage benchmark task.
Inputs: level, domain, user role, analyst style, columns, visible data preview,
reasoning graph, canonical skill registry, and 80/20 visible/held-out split
metadata.

Hard rules:
1. Cover every reasoning-graph node in at least one agent turn.
2. Use only canonical skill names from the registry.
3. Start with a user turn and preserve open-loop evaluability: user turns may
   reference prior concrete state but must not depend on the scripted agent's
   opinions.
4. Do not leak held-out rows, held-out values, split boundary indices, total
   length, or held-out row counts.
5. Every analytical agent turn must include reference_code with
   "# SKILL_USED: <skill_name>" and print every digit-bearing narrative claim.
6. L3 must include synthesis gold; L4 must end with decision_json.
7. Output exactly the requested number of turns as JSON only.
\end{verbatim}
}

\input{table/appendix_table/appendix_b/P1/5_data_generation}

\subsection{P2: Reproducibility Audit}

P2 runs 46 deterministic checks grouped into 10 categories, all auditable from the task output. A task is held back from P3 until every P2 check passes. Table \ref{tab:p2_checklist} describes each check its subchecks.

\input{table/appendix_table/appendix_b/P2/1_reproduce_checklist}

\subsection{P3: LLM Cross-Family Review}

P3 runs an anti-bias LLM audit. The model used in P3 is from a different family than the S5 generator: tasks generated by Claude Opus~4.6 are reviewed by a GPT-family model (default GPT-5.4 used in this benchmark). P3 evaluates 4 critical checklist items; any single failure rejects the task back to S5 for regeneration and recheck from P2. Items 11–14, illustrated in the prompt excerpt below, align with the four most common P3 failure modes we observed during construction (factual drift, test-set leakage, training-data memorization, open-loop evaluability).

\paragraph{P3 LLM prompt excerpt.}
{\small
\begin{verbatim}
You are an independent benchmark reviewer. Be adversarial and precise.
Cite turn_id, field, or quoted text for every failure. Fail only CRITICAL
evaluation-contract defects.

Apply these checks:
11 domain_knowledge_correct:
   Analytic claims must be consistent with reference stdout and domain knowledge.
12 no_heldout_leakage:
   No dialogue, reference-code comment, or agent-visible text may reveal
   held-out values, held-out row counts, split boundary integers, or
   scorer-only gold. Scorer gold hidden from the evaluated agent is allowed.
13 no_training_memorization_artefact:
   The task must not copy known Kaggle/forum/textbook/canonical analyses.
14 open_loop_evaluable:
   Each user turn must be answerable from the visible CSV, task context, and
   prior user turns only; it must not depend on scripted agent opinions.

Return JSON:
{
  "items": [{"id": "...", "pass": 1 or 0, "note": "...",
             "failing_turn_ids": [..]}],
  "overall_recommendation": "auto_approve" or "reject",
  "reviewer_notes": "..."
}
\end{verbatim}
}

\subsection{P4: Assigned Human Review in the Annotation Dashboard}

\begin{figure}[htbp]
\centering
\includegraphics[width=0.95\linewidth]{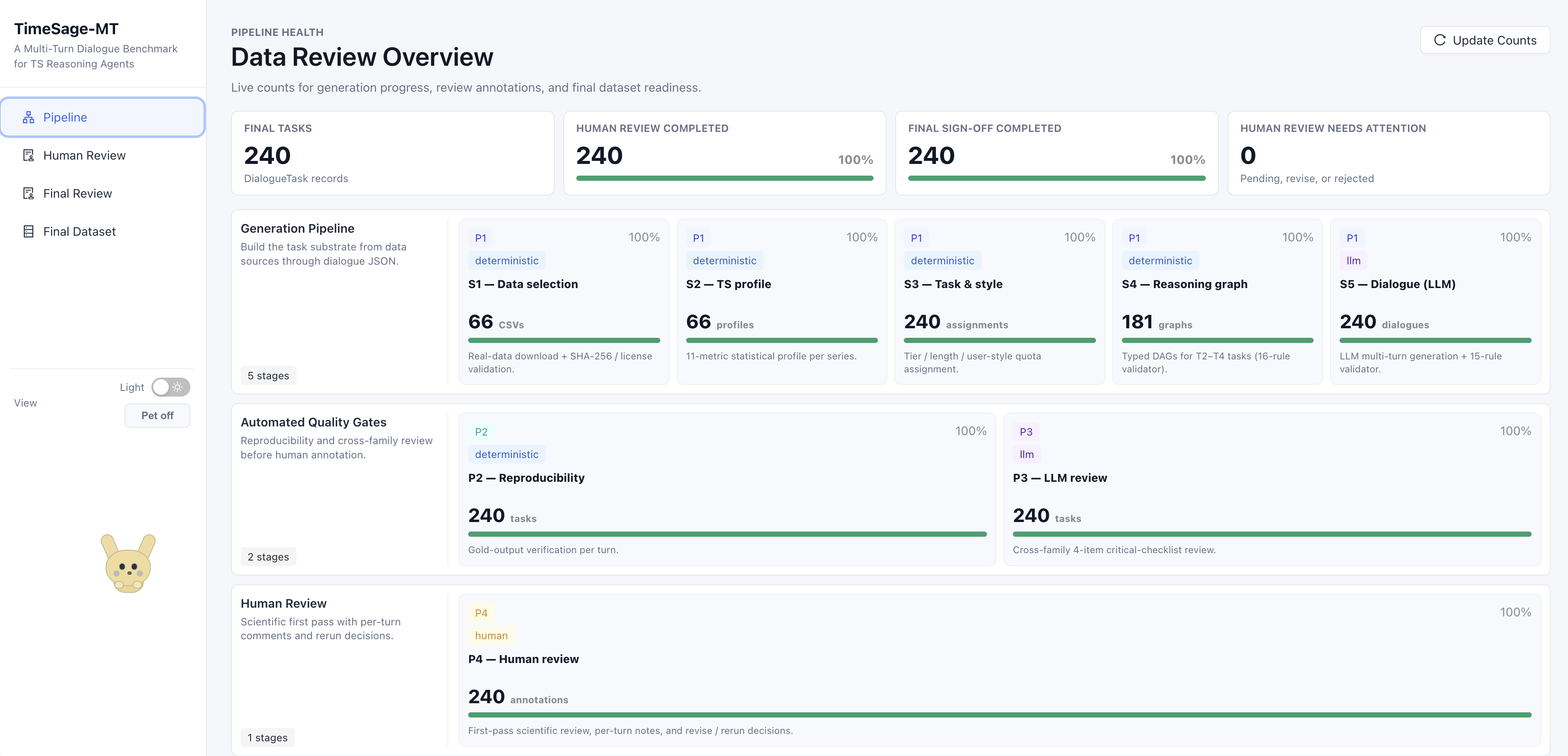}
\caption{Overview of TimeSage-MT benchmark construction dashboard. The dashboard summarizes phase-level progress across data selection, profiling, task assignment, reasoning-graph construction, dialogue generation, automated checks, human review, final review, and final dataset release.}
\label{fig:data_pipeline_overview_dashboard}
\end{figure}

\begin{figure}[htbp]
\centering
\includegraphics[width=0.95\linewidth]{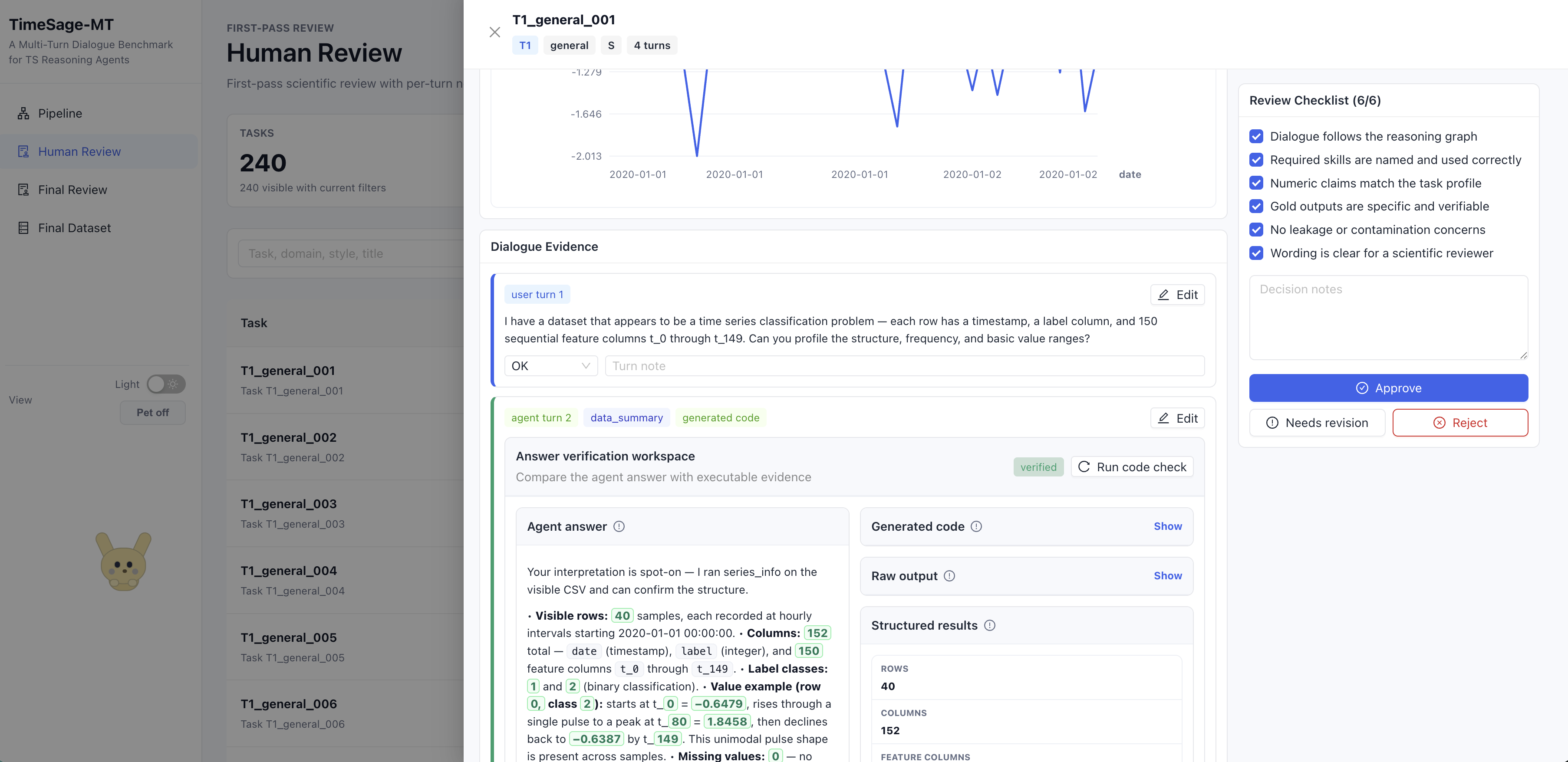}
\caption{Human annotation and review dashboard used in P4. The interface supports task-level review through the task queue, dialogue view, reasoning graph, gold fields, turn-level annotations, and approve/revise/reject controls.}
\label{fig:human_review_annotation_dashboard}
\end{figure}

P4 is human review with dashboard support (Figure \ref{fig:data_pipeline_overview_dashboard} and \ref{fig:human_review_annotation_dashboard}). Each candidate task is assigned to one reviewer. The reviewer inspects the source data, profile, visible CSV preview, reasoning graph, dialogue, reference code, gold fields, P2 report, and P3 report. Reviewers can mark individual turns as OK, flagged, or requiring revision; edit task output (in JSON format); trigger per-task S5 regeneration; approve; reject; or request revision. A task is eligible for release only after its assigned human reviewer approves it and it survives the later P5 reproducibility checks. Table \ref{tab:human_review_checklist} below shows the human review checklist that helps and guides the reviewing process.

\input{table/appendix_table/appendix_b/P4/1_human_review_checklist}

\subsection{P5: Final Reproducibility and Release TimeSage-MT}
\label{appendix_b_p5}

P5 is the release gate. Three things happen in sequence (Table \ref{tab:final_release_checklist}): (i) a fresh P2 audit is re-run end-to-end to catch any drift introduced by P4 edits; (ii) all reference code is re-executed inside a clean sandbox container and the recorded output is verified to match the gold target within tolerance; and (iii) a human reviewer inspects any task
flagged by P5 spot-checks. Tasks that pass P5 are written to final data and become release-frozen. Failures re-enter the fix~$\to$~P2~$\to$~P3 cycle. The dashboard \ref{fig:data_pipeline_overview_dashboard} records P5 status alongside P4, so reviewers see both labels at once and the release log is auditable.

\input{table/appendix_table/appendix_b/P5/1_final_check}

\section{TimeSage-MT Evaluation Protocol}
\label{appendix_c}

TimeSage-MT produces two parallel score tracks per task. \textbf{Outcome} scores measure whether the agent produced correct answers. \textbf{Capability} scores measure whether the agent exhibited the intended analytical behavior along the way. Both views operate turn-by-turn when possible and then aggregate to the task, level, and system
levels. Capability has nine axes. Most sub-checks are deterministic; only four (analytical quality, decision quality, C7 Decide, C9 Ground) use an LLM judge. Every LLM-judged sub-check is content-addressed by a SHA-256 cache key so re-evaluations on the same content return cached scores deterministically. 

\subsection{Outcome Evaluation}
\label{app:outcome_evaluation}

The Outcome track grades each task along up to four dimensions: \textbf{numerical accuracy}, \textbf{key findings coverage}, \textbf{decision accuracy}, and \textbf{narrative match}. It compares the agent's response, code output, and final structured output to task gold. Rule-based components are preferred whenever the target is numerical, categorical, structured, or directly extractable. LLM judges are
used only for narrative quality and decision quality, where there may be many valid answers. Metric applicability is level-conditional: numerical, factual, code, and analytical-quality scores are computed whenever the corresponding gold field or response type exists for an L1--L4 task, whereas decision quality is computed only for L4 tasks because only L4 ends in a structured decision node. Decision quality is therefore not counted as zero for L1--L3; it is marked not applicable and excluded from those level-specific denominators.

Let $M(t)$ be the set of non-null outcome components for task $t$.  Missing or
not-applicable components are dropped rather than filled with zero.  The
outcome score is
\[
  \mathrm{Outcome}(t) =
  \begin{cases}
  |M(t)|^{-1}\sum_{m\in M(t)} s_m(t), & |M(t)| > 0,\\
  \mathrm{N/A}, & |M(t)| = 0.
  \end{cases}
\]
System-level results report macro-averages over scorable tasks, with additional breakdowns by level, domain, length bucket, and user style. Table \ref{tab:outcome_metrics} records outcome metrics and scoring rules.

\input{table/appendix_table/appendix_c/1_outcome_evaluation}

\paragraph{Analytical-quality judge prompt excerpt.}
The analytical-quality judge is applied to free-form analytical responses in L1--L4 when such a response is scorable. It compares the agent response against the actual data, reference stdout, and one valid reference answer. The excerpted instruction is:

{\small
\begin{verbatim}
You are evaluating whether an AI agent's time series analysis is correct and
insightful, judged against the actual data -- NOT against a specific reference
answer.

Score on three axes in [0,1]:
- factual_correctness
- analytical_depth
- conclusion_soundness

The example answer is one valid response, not the correct response. Penalize
only factual errors, fabricated claims, or unsupported conclusions. Return the
arithmetic mean as a single number.
\end{verbatim}
}

\paragraph{Decision-making judge prompt excerpt.}
The decision-making judge is applied only to L4 tasks, because only L4 contains
a final structured decision node.  It uses decision-specific axes and compares
the final decision against the actual data and reference stdout, while treating
the reference decision as one valid decision rather than the only correct
decision.  The excerpted instruction is:

{\small
\begin{verbatim}
You are evaluating the quality of an AI agent's decision on a time series
analysis task, judged against the actual data -- NOT against a specific
reference decision.

Score on three axes in [0,1]:
- decision_validity
- reasoning_quality
- completeness

The reference decision is one valid decision, not the only correct decision.
Return the arithmetic mean as a single number.
\end{verbatim}
}

For readability, we report shortened excerpts of the two outcome LLM judge prompts rather than the full implementation templates, which can be accessed in the codebase.

\subsection{Capability Evaluation}
\label{app:capability_evaluation}

Capability evaluation scores each agent on the nine capabilities C1--C9 using
capability-specific subchecks.  Each subcheck emits a score in [0,1] or N/A.
C1--C6 and C8 are fully deterministic: their subchecks inspect the agent's
structured trace against the task's reasoning graph, gold fields, and expected
intermediate outputs.  C7 Decide and C9 Ground are hybrid: deterministic
subchecks score structural, citation, and domain-term evidence, while LLM judge
subchecks score cross-turn coherence and domain-correctness judgments that
resist purely rule-based capture. 

For a capability $c$ on task $t$, let $S_c(t)$ be the set of fired subchecks. The capability score is 
\[
  \mathrm{Cap}_c(t) =
  \begin{cases}
  |S_c(t)|^{-1}\sum_{s\in S_c(t)} s(t), & |S_c(t)| > 0,\\
  \mathrm{N/A}, & |S_c(t)| = 0.
  \end{cases}
\]
The overall capability score for a task is the mean of non-null capability scores.  Capability macro-averages are then computed over tasks where the capability is applicable.  In total, the implementation exposes 28 possible capability subchecks; 26 are deterministic or local-embedding checks and two are LLM judge checks. Table \ref{tab:capability_subchecks} illustrates the capability subchecks and aggregation. 

\input{table/appendix_table/appendix_c/2_capability_evaluation}

\paragraph{Capability LLM judge prompt excerpt.}
\label{app:capability_LLM_judge}

C7 uses the following compact coherence prompt:

{\small
\begin{verbatim}
You are evaluating whether an AI agent's conclusion is consistent with its
analysis.

Agent's analysis results: {skill_results_summary}
Agent's final conclusion: {agent_synthesis}

Is the conclusion consistent with the analysis?
Analysis supports conclusion -> 1; contradicts -> 0; partial support -> 0.5.
Output only the number.
\end{verbatim}
}

C9 uses a domain expert prompt:

{\small
\begin{verbatim}
You are a {domain} domain expert.

Agent's analysis results: {analysis_summary}
Agent's domain interpretation: {agent_domain_reasoning}

Is the domain reasoning sound?
Domain logic correct -> 1; clear domain errors -> 0; partial -> 0.5.
Output only the number.
\end{verbatim}
}

The judge infrastructure caches prompts by task ID, agent ID, capability, judge model, and prompt hash.  The default configuration uses a single cross-family judge; two-judge averaging is available when the judge pool is configured with two different model families.

For readability, we report shortened excerpts. The full prompts can be accessed in the codebase.

\subsection{Why We Use Joint Outcome--Capability Evaluation}
TimeSage-MT uses a joint outcome--capability protocol because multi-turn time series analysis is not fully characterized by final-answer correctness alone. Outcome scores measure whether the agent reached a correct numerical, categorical, narrative, or decision output. Capability scores measure whether the agent used the intended analytical process: selecting appropriate skills, tracking dependencies across turns, respecting the visible/held-out split, handling corrections, synthesizing evidence, and grounding claims in domain knowledge. Reporting both views separates answer quality from process quality, which is important when different agentic systems can arrive at similar final answers through very different reasoning paths. 

The two views also constrain each other. Outcome scoring anchors the evaluation to task correctness, so an agent is not rewarded merely for producing a plausible analytical trace. Capability scoring explains how the outcome was obtained, so a correct answer can be distinguished from shortcutting, leakage, or lucky agreement with the gold target. Combining them therefore improves diagnostic value: it identifies whether failures come from wrong final calculations, missing skills, poor cross-turn state tracking, weak grounding, or decision synthesis errors.

The protocol therefore treats outcome and capability as complementary axes. For each task, the benchmark records an outcome score $O(t)$ and a capability score $K(t)$ over applicable capabilities. The primary analysis reports both scores and their breakdowns by level, domain, length bucket, and user style. Researchers using our benchmark in the future may additionally expose a transparent composite such as
$\frac{1}{2}O(t)+\frac{1}{2}K(t)$ for ranking convenience, but in this work, we keep the two components visible so that a high final score cannot hide weak reasoning behavior. Table~\ref{tab:joint_outcome_capability} interprets the
four qualitative regimes induced by the two evaluation axes.

\input{table/appendix_table/appendix_c/3_joint}

\section{Evaluated Agentic Systems and Public Interfaces}
\label{appendix_d}

This section describes the four evaluated systems and the public interfaces released with TimeSage-MT. All systems receive the same user turns and visible data contract. They differ only in tool access, skill access, and orchestration. Table \ref{tab:agentic_systems} demonstrates the comparison of the evaluated agent systems. 

\input{table/appendix_table/appendix_d/1_comparison_agentic_system}

\subsection{System A: Direct Answering}

The direct-answer baseline receives the visible data summary or CSV text and the current user turn. It has no tools, no Python execution, no skill library, and no access to scripted agent traces. It measures how much time series reasoning can be obtained from the base model and prompt alone.

{\small
\begin{verbatim}
System prompt excerpt:
You are a time series analyst. A user is going to ask you a series of questions
about a dataset. For each question, think about what analysis is appropriate
and produce the answer directly. If numerical output is required, put a
parseable JSON array on the last line. If a structured decision is required,
return a JSON object with decision, rationale, and confidence. You do NOT have
access to tools or skills.
\end{verbatim}
}

\subsection{System B: Code-Enabled Reasoning}

The Code-Enabled Reasoning baseline receives the same open-loop user turns, but can write and execute Python in a sandbox. Each turn follows a two-call protocol: the model first emits code, the sandbox executes it, and the model then answers using stdout and stderr. We also introspect the generated code to recover method choices and parameters for capability scoring.

{\small
\begin{verbatim}
Code-generation prompt excerpt:
You are a time series analyst with access to a Python sandbox. Output only one
Python fenced code block. Load the dataset path provided in the prompt. You may
use pandas, numpy, scipy, sklearn, statsmodels, ruptures, and matplotlib. Print
the answer, and if numerical output is required, print a JSON array or object
on the last line. Never overwrite the input data.

Explanation prompt excerpt:
The sandbox ran your code. Here is stdout and stderr. Now answer the user's
most recent question using actual numbers from the execution and prior turns.
\end{verbatim}
}

\subsection{System C: Skill-Guided Code Reasoning}

This baseline adds the released TimeSage skill library to Code-Enabled Reasoning. The model still writes one code block per turn, but can call canonical time series routines through a compact API. This setting isolates the effect of the skill library from the full TimeSage orchestration harness.

{\small
\begin{verbatim}
Skill API prompt excerpt:
You have access to a curated time series skill library:
from skill_library.runner import call_skill, describe_skill, list_skills

call_skill(name, data, *, target_column=None, datetime_column=None, **params)
returns a SkillResult with outputs, metrics, artifacts, and diagnostics.

Available skills are grouped by category. Choose canonical skill names and
print the final numerical or structured answer on the last line.
\end{verbatim}
}

\subsection{System D: TimeSage}

TimeSage is a stateful agentic harness for time series analysis. It maintains multi-turn session state, plans analytical steps, invokes tools and skills, validates results, and produces a final synthesis or decision. The evaluated harness receives only the user turns, preserving the open-loop benchmark setting. Its pipeline combines deterministic profiling, LLM-based planning, skill/tool execution with retries, rule-based result checks, and LLM-based reporting.

{\small
\begin{verbatim}
System prompt excerpt:
You are Time_Sage, a specialized AI assistant for time series analysis.
You have access to a broad library of analytical skills. For each task, select
the most appropriate skill(s), configure them carefully, chain skills when
needed, validate results, and provide grounded interpretations.

Planning prompt excerpt:
Understand the data first, select suitable methods, validate/evaluate outputs,
and provide interpretations. Terminate when the user's analytical goal has been
completed.
\end{verbatim}
}

\subsection{Public Leaderboard, Methodology Dashboard, and Agentic Platform}

\begin{figure}[htbp]
\centering
\includegraphics[width=0.95\linewidth]{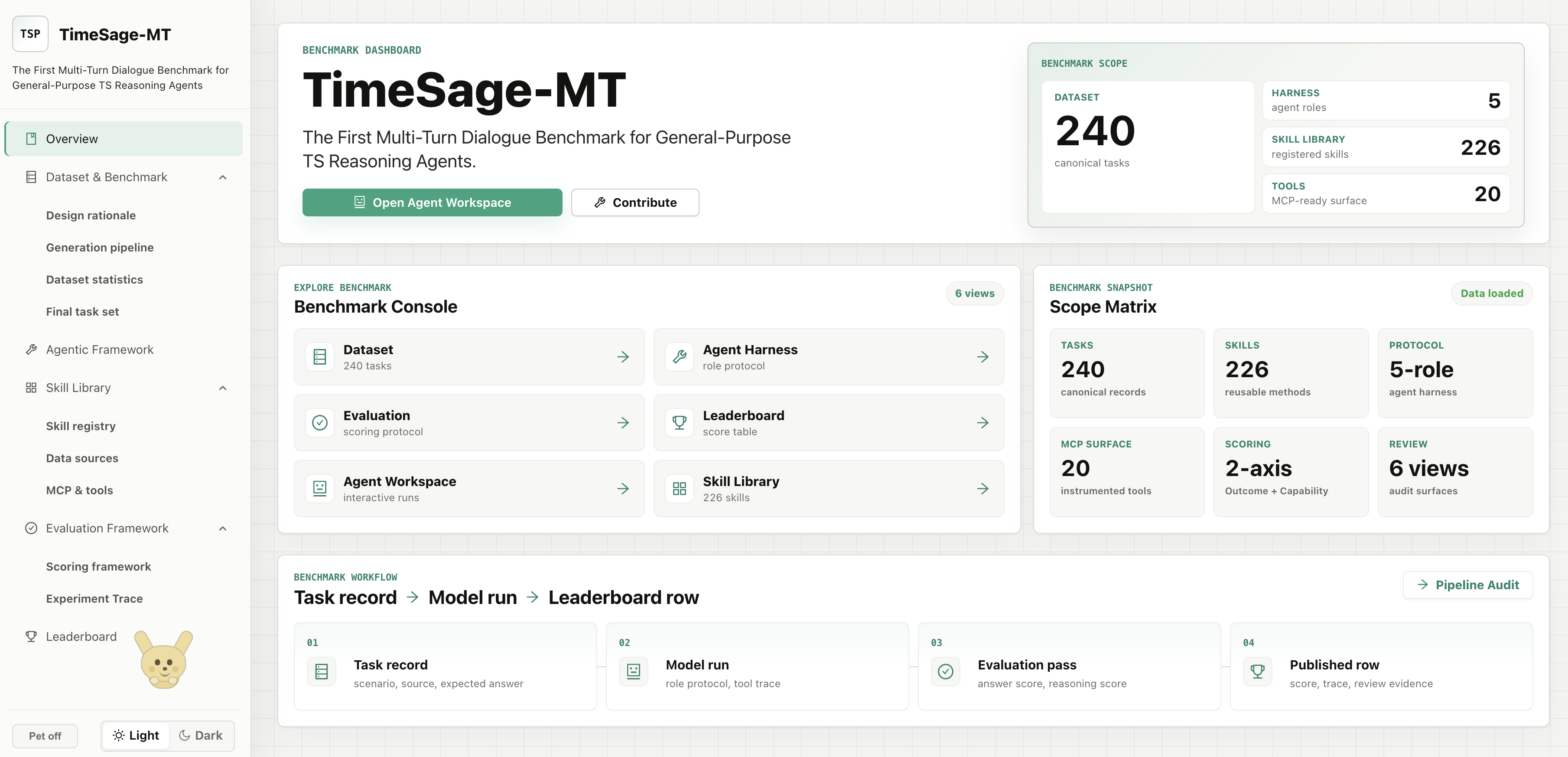}
\caption{Leaderboard and methodology dashboard shipped with the TimeSage-MT release. The dashboard summarizes benchmark results, outcome and capability breakdowns, paper overview, benchmark methodology, and open-source release notes.}
\label{fig:leaderboard_methodology_dashboard}
\end{figure}

\begin{figure}[htbp]
\centering
\includegraphics[width=0.95\linewidth]{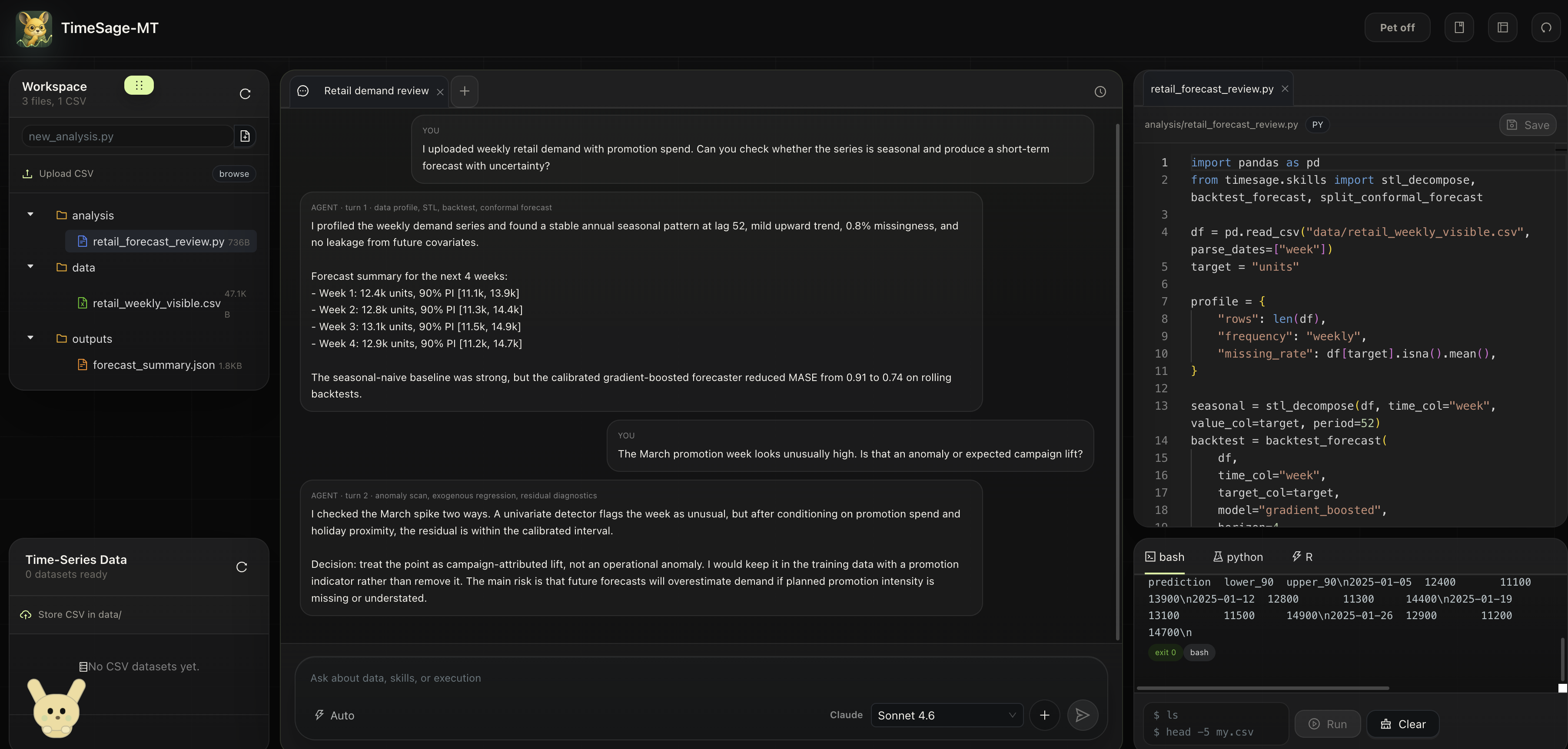}
\caption{Interactive agentic platform for multi-turn time series analysis. The interface displays a conversation, selected workspace files, executable analysis code, terminal output, dataset controls, and model/tool controls used to support agent-driven analysis.}
\label{fig:agentic_platform_dashboard}
\end{figure}

\begin{figure}[htbp]
\centering
\includegraphics[width=0.95\linewidth]{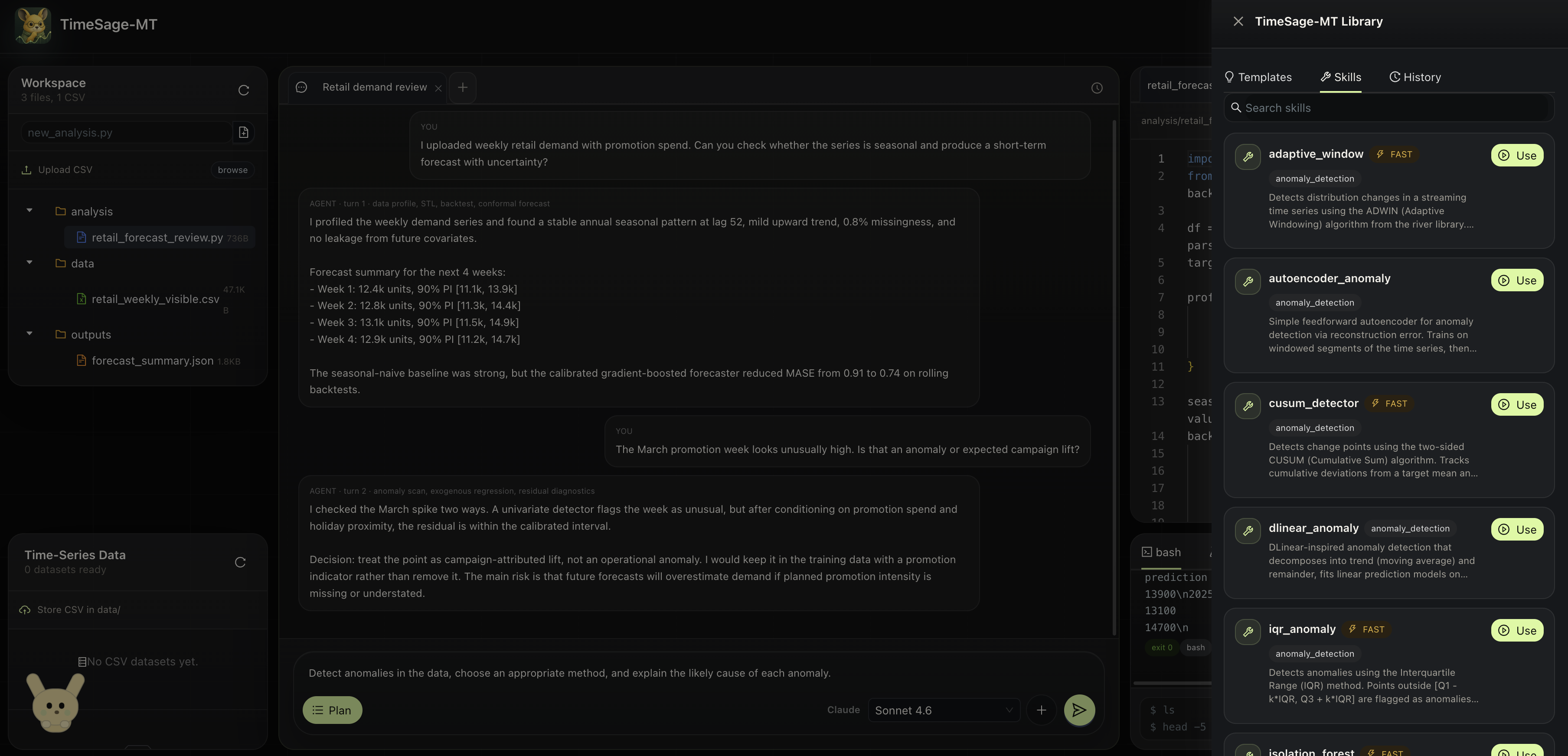}
\caption{Interactive agentic platform with TimeSage-MT skill library panel open. This allows users to inspect and choose available time series skills during multi-turn analysis, including skill categories, searchable skill cards, scientific-use descriptions, evidence accepted, and evidence returned.}
\label{fig:skill_library_dashboard}
\end{figure}

We release the dashboard and platform through the project website and the code repository. The open-source dashboard (illustrated in Figure \ref{fig:leaderboard_methodology_dashboard}) covers leaderboard results, paper overview, and benchmark methodology, and makes aggregate results inspectable by level, capability, outcome metric, domain, style, and agent system. The leaderboard reports the four evaluated systems: Direct Answering, Code-Enabled Reasoning, Skill-Guided Code Reasoning, and TimeSage. We also publicly release the interactive agentic platform (illustrated in Figure \ref{fig:agentic_platform_dashboard} and \ref{fig:skill_library_dashboard}).  Users can connect different LLM or agents, upload data, build their own agentic system to test and run multi-turn time series analysis conversations using the same general harness interface.

\section{TimeSage Skill Library and Harness Engineering}
\label{appendix_e}

TimeSage separates three components that are often entangled in agent benchmarks: method coverage, orchestration, and evaluation. The skill library provides a stable vocabulary of time series operations. The harness controls multi-turn state, data visibility, tool execution, retries, and logging. The evaluator then scores whether an agent selected, configured, chained, corrected, calibrated, and grounded these operations correctly. This design makes failures more interpretable, since errors can be traced to specific capabilities rather than to a single aggregate answer score.

\subsection{Released Skill Library}

The released library contains 226 skills across 36 categories. Each skill has a completed definition card that specifies the name, category, inputs, outputs, parameters, dependencies, artifacts, diagnostics, and usage notes. We make the cards publicly available and easy to inspect through the methodology dashboard, so that other researchers can reuse the library, extend it with new methods, or plug the same skill definitions into their own agents. Table~\ref{tab:skill_inventory} summarizes the released skill inventory by category and shows the coverage breadth of the library.

\input{table/appendix_table/appendix_e/1_skill_library}

\begin{figure*}[t]
\centering
  \includegraphics[width=0.98\linewidth]{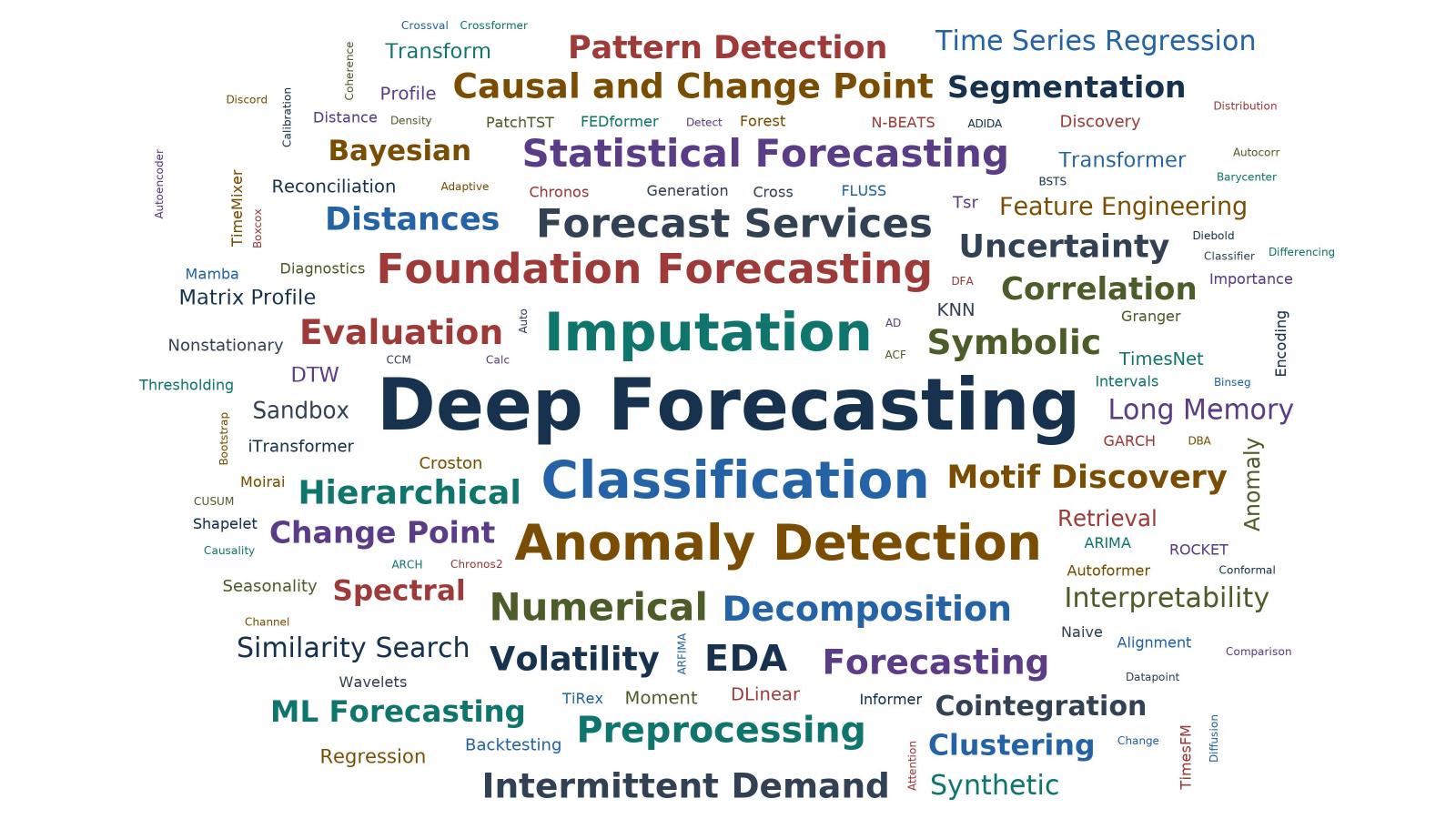}
  \caption{Taxonomy word cloud for the TimeSage skill library. Terms are
  generated from the canonical names and categories of the 226 released skills;
  larger terms indicate greater representation.}
  \label{fig:skill_library}
\end{figure*}

\paragraph{Skill taxonomy visualization.}
For readability, we present the skill library as a taxonomy word cloud. Figure \ref{fig:skill_library} is generated directly from the 226 skill cards: category names and skill names are weighted by their frequency, so larger terms indicate broader coverage. The complete skills remain available in the codebase and through the methodology dashboard.

\subsection{Tools, MCP Interface, and Data Sources}

The harness exposes direct tools, MCP-compatible wrappers, and a curated data-source registry because the benchmark is designed to evaluate agents as working analytical systems rather than as text-only responders. Direct tools provide controlled execution, file access, visualization, and logging inside the benchmark harness. The MCP interface lets external agents discover skills and tools through schemas rather than benchmark-specific imports. The skill adapter converts each definition card into a callable tool schema, loads CSV inputs, executes the selected skill, and returns structured outputs, metrics, diagnostics, and artifacts.  Table~\ref{tab:tool_mcp_inventory} summarizes the tool and MCP inventory and states the purpose of each interface group.

\input{table/appendix_table/appendix_e/2_tool}

\begin{figure*}[t]
\centering
  \includegraphics[width=0.98\linewidth]{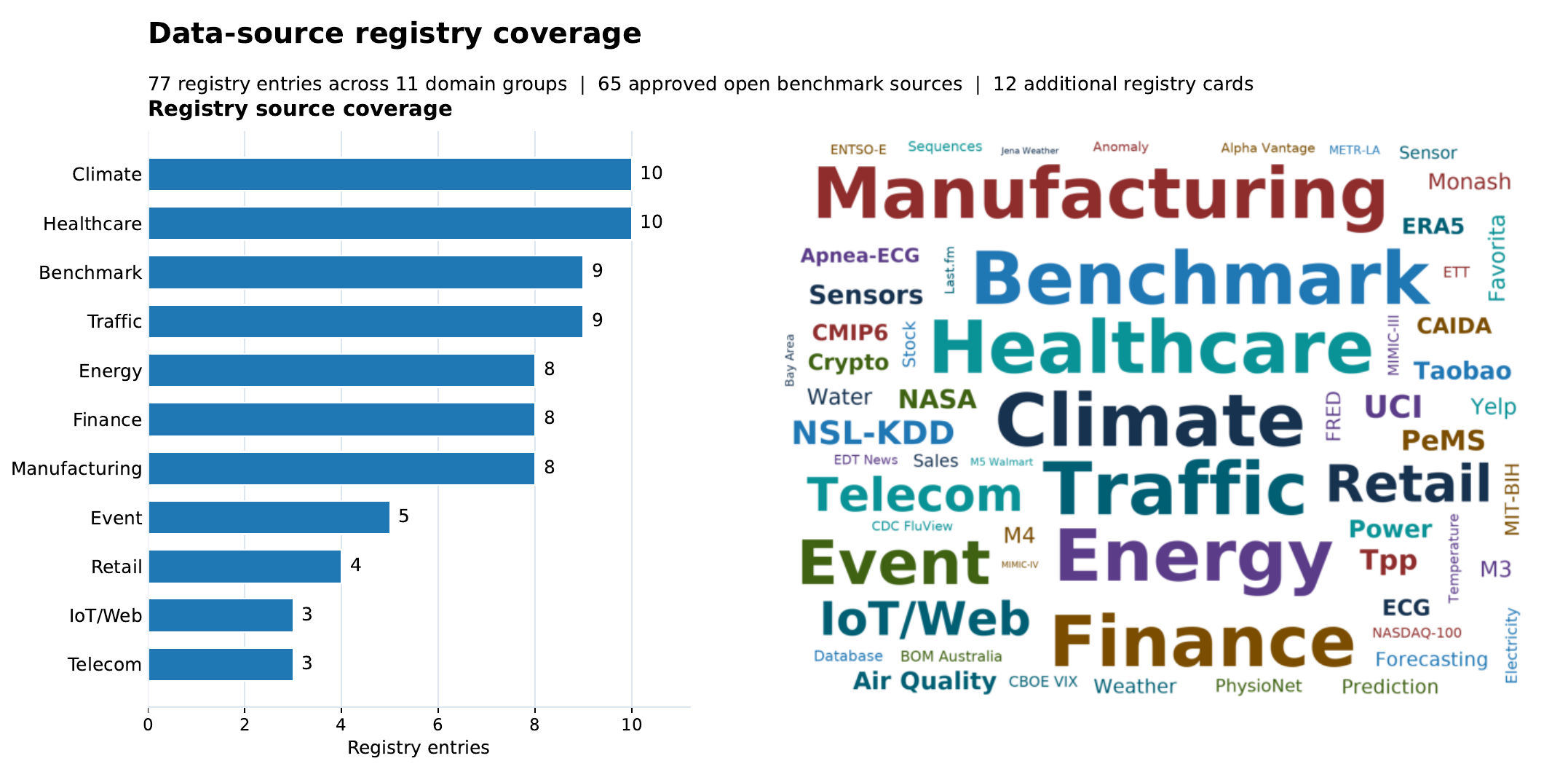}
  \caption{Data-source registry coverage.  The left panel reports the number of
  registry entries in each domain, stacked by access policy.  The right panel
  summarizes source names, provider families, and domain terms as a word cloud
  generated from the 77-entry data-source registry.}
  \label{fig:data_source_coverage}
\end{figure*}

The broader data-source registry contains 77 named sources across 12 domains. It is built to provide realistic time series material for task generation, domain grounding for synthesis and decision tasks, reproducible loading metadata for external agents, and a controlled source pool for license, parseability, length, and contamination checks. The benchmark construction pipeline uses a whitelisted subset of 65 real-world sources after these checks. Figure~\ref{fig:data_source_coverage} shows the registry coverage by domain and access status and summarizes the source taxonomy as a word cloud. The complete source list is included in the codebase and can be exposed through the methodology dashboard.

\subsection{Harness Engineering}

The harness is built around reproducibility, traceability, and fair comparison across agent types. All evaluated systems receive the same user turns and the same visible data contract. Agents that can execute code or skills do so in logged environments, and evaluators recover behavior from traces, stdout, structured outputs, and code introspection rather than from final text alone. Table \ref{tab:harness_components} illustrates each component's engineering role and rationale.  

\input{table/appendix_table/appendix_e/3_harness}

This resulting design makes TimeSage-MT both a benchmark and an engineering artifact. In the future, researchers can evaluate their own agents on the released tasks, inspect capability-level failures, reuse the 226-skill library through direct imports or MCP, and extend the corpus under the same construction and evaluation contracts.

\section{Further Results}
\label{appendix_f}

This section reports additional leaderboard details underlying the main results in Section~\ref{main_leaderboard_performance}. We provide complete breakdowns by difficulty tier, outcome dimension, fine-grained capability, and the paired Skill-Guided versus TimeSage comparison.

\subsection{Performance Across Difficulty Tiers}

\input{table/appendix_table/appendix_f_result/1_performance_tier}

Table~\ref{tab:perf-tier} reports overall scores for each model across the four difficulty tiers. The full tier-wise results show that all models achieve their highest scores on L1 and drop at L2, with smaller changes from L2 to L4. Across the leaderboard, overall scores remain within a narrow range, from 35.21 for Gemini-3-Flash-Preview to 40.50 for GLM-5.1. These detailed numbers support the tier-level trends summarized in Figure~\ref{Figure_6}.

\subsection{Performance Across Outcome Dimensions}

\input{table/appendix_table/appendix_f_result/2_performance_outdim}

Table~\ref{tab:perf-outdim} reports the five outcome dimensions for each model. Code Correctness is the highest-scoring dimension across models, ranging from 71.7 to 78.5, while Numerical Accuracy is the lowest, ranging from 15.3 to 18.2. The table also shows that the leading model differs by dimension: GPT-5.3-Codex obtains the highest Analytical Quality, MiniMax-M2.7 the highest Decision Making, GLM-5.1 the highest Numerical Accuracy, and Claude-Sonnet-4.6 the highest Factual Verification and Code Correctness. These results provide the full numerical decomposition behind Figure~\ref{Figure_7:radar}.

\subsection{Performance Across Capabilities}

\input{table/appendix_table/appendix_f_result/3_performance_capability}

Table~\ref{tab:perf-capability} gives the nine-capability breakdown. C4 Remember and C5 Chain are consistently high across models, with scores of 80.5--88.8 and 94.1--99.8, respectively. In contrast, C1 Read and C3 Configure are consistently low, with scores of 22.2--27.5 and 19.3--20.1. Decision-related capabilities fall in the middle range: C7 Decide reaches 64.5--77.1, C8 Calibrate remains at 42.1--44.6, and C9 Ground ranges from 55.1 to 70.1. The capability scores correspond to the heatmap in Figure~\ref{Figure_7:heatmap}.

\subsection{Direct Reasoning vs. Code-Enabled Reasoning}

\input{table/appendix_table/appendix_f_result/4_direct_vs_code}

Table~\ref{tab:direct-vs-code} reports the paired comparison between
Direct Answering and Code-Enabled Reasoning on the 100-task subset used
in Section~\ref{section_4.4_key_observations}. For GPT-5.3-Codex,
Code-Enabled Reasoning increases the overall score by 24.46 points.
The largest increases are in Code Correctness ($+78.63$), where Direct
Answering scores zero by construction because it produces no executable
evidence, followed by Decision Making ($+15.77$), Factual Verification
($+7.92$), and Numerical Accuracy ($+6.57$); Analytical Quality also
improves slightly ($+3.28$). For Qwen-3.5-122B-A10B, Code-Enabled
Reasoning increases the overall score by 19.57 points. The largest
gains are again in Code Correctness ($+73.67$) and Decision Making
($+23.12$), with smaller positive shifts in Analytical Quality
($+5.32$) and Numerical Accuracy ($+1.56$); the only negative shift is
Factual Verification ($-5.34$), suggesting that Qwen's free-form direct
answers report profile attributes more readily than its code-driven
narratives. The tier-wise rows further show that the Code-Enabled
gains are present across all four tiers for both backbones, confirming
that execution-based grounding helps regardless of difficulty level.
These results provide the full numerical details behind
Observation~\#1.
 
The tier-wise outcome decomposition shows that Code-Enabled Reasoning
delivers its largest benefits on evidence-oriented dimensions and at
the entry tiers, where execution is enough to lift previously
unsupported answers. For GPT-5.3-Codex, Numerical Accuracy improves at
every tier, with the largest gain at T1 ($+12.00$), and Factual
Verification improves at every tier, peaking at T2 ($+10.72$).
Analytical Quality is uniformly higher except for a small dip at T2
($-1.20$), and Decision Making improves on the only tier where it is
evaluated (T4, $+15.77$). For Qwen-3.5-122B-A10B, Numerical Accuracy
and Decision Making (T4 only, $+23.12$) improve at every applicable
tier, while Factual Verification declines at every tier, with the
largest drop at T4 ($-8.43$); Analytical Quality improves at three of
four tiers, with the largest gain at T3 ($+7.60$) and a near-zero
change at T2 ($-0.08$). Overall, the across-tier results show that
moving from Direct Answering to Code-Enabled Reasoning produces a
large, broadly distributed gain driven by execution-based grounding,
but it does not uniformly improve narrative-aligned profile
verification on every backbone, motivating the additional structure
introduced in subsequent settings.

\subsection{With Skill vs. Without Skill}

\input{table/appendix_table/appendix_f_result/5_skill_vs_noskill}

Table~\ref{tab:code-vs-skill} reports the paired comparison between
Code-Enabled Reasoning and Skill-Guided Code Reasoning on the same
100-task subset. For GPT-5.3-Codex, the skill library increases the
overall score by 5.52 points. The largest increase is in Analytical
Quality ($+25.72$), followed by Decision Making ($+12.32$), while
Numerical Accuracy ($-0.45$), Factual Verification ($-1.18$), and Code
Correctness ($-1.34$) decrease slightly. For Qwen-3.5-122B-A10B, the
skill library increases the overall score by only 0.38 points. Code
Correctness improves by 2.60 points, but Numerical Accuracy ($-0.18$),
Factual Verification ($-0.56$), and Analytical Quality ($0.00$) are
essentially unchanged, and Decision Making changes by only 0.12. The
tier-wise rows further show that GPT-5.3-Codex receives a substantial
Analytical Quality boost from the skill library at every tier, whereas
Qwen-3.5-122B-A10B shows near-zero changes on most dimensions across
all tiers. These results provide the full numerical details behind
Observation~\#2.
 
The tier-wise outcome decomposition shows that the skill library mainly
improves expressive and decision-oriented dimensions for the stronger
backbone, while leaving the weaker backbone largely unaffected. For
GPT-5.3-Codex, Analytical Quality improves at every tier, with the
largest gains at T3 ($+27.46$) and T4 ($+27.12$) and a smaller but
still substantial gain at T2 ($+20.59$). Decision Making (evaluated
only at T4) improves by $+12.32$, while Numerical Accuracy and Factual
Verification show small mixed shifts (e.g., FV $-3.45$ at T4, NA
$-1.31$ at T4). For Qwen-3.5-122B-A10B, the across-tier picture is
mostly flat: Analytical Quality changes by less than $\pm0.5$ points at
T1 and T3 ($-0.43$ and $-0.53$ respectively) and by $+1.18$ at T2,
while Numerical Accuracy and Factual Verification stay within a
$\pm1.5$-point band at every tier. Code Correctness improves at three
of four tiers (peaking at T3, $+4.20$) but decreases slightly at T1
($-0.12$). Overall, the across-tier results show that adding a
predefined skill library on top of Code-Enabled Reasoning yields large,
expressiveness-driven gains for a backbone that can flexibly invoke and
integrate skills, but produces only marginal changes when the backbone
does not exploit the skill interface effectively, motivating the
further structural orchestration introduced in TimeSage.

\subsection{Skill-Guided Reasoning vs. TimeSage}

\input{table/appendix_table/appendix_f_result/6_skill_vs_timesage}

Table~\ref{tab:skill-vs-timesage} reports the paired comparison between Skill-Guided Code Reasoning and the full TimeSage pipeline on the 100-task subset used in Section~\ref{section_4.4_key_observations}. For GPT-5.3-Codex, TimeSage decreases the overall score by 3.45 points. The largest decreases are in Analytical Quality ($-19.84$) and Decision Making ($-24.03$), while Factual Verification increases by 7.85 and Numerical Accuracy by 2.21. For Qwen-3.5-122B-A10B, TimeSage increases the overall score by 0.65 points. The largest gains are in Numerical Accuracy ($+6.57$), Analytical Quality ($+4.75$), and Factual Verification ($+2.67$), while Decision Making decreases by 28.20 and Code Correctness by 5.11. The tier-wise rows further show that Qwen's Numerical Accuracy improves at all four tiers, whereas GPT-5.3-Codex shows larger losses on open-ended and decision-oriented dimensions. These results provide the full numerical details behind Figure~\ref{Figure_8}.

The tier-wise outcome decomposition shows that TimeSage mainly improves evidence-oriented dimensions while hurting expressive or decision-oriented ones. For GPT-5.3-Codex, Numerical Accuracy and Factual Verification improve at every tier, with the largest Factual Verification gain at T4 ($+14.88$). However, Analytical Quality drops sharply at T1 ($-26.50$) and T3 ($-27.75$), and Decision Making drops at T4 ($-24.03$), leading to a lower overall score. For Qwen-3.5-122B-A10B, Numerical Accuracy improves consistently across all tiers, especially at T1 ($+10.40$), and both Factual Verification and Analytical Quality improve from T2 to T4. These gains are offset by lower Code Correctness at every tier and a large T4 Decision Making drop ($-28.20$). Overall, the across-tier results show that TimeSage strengthens numerical grounding and verification, but can weaken code generation, free-form analysis, and final decision expression.

\section{Intended Use and Scope of Validity for TimeSage-MT}
\label{appendix_g}
TimeSage-MT is intended to evaluate time series analysis as an auditable multi-turn workflow rather than as a single prediction, a single prompt, or a free-form conversation. The benchmark should be used to compare time series agents under a fixed visibility contract, diagnose failures by level and capability, and study how models use code, tools, skills, memory, uncertainty estimates, and domain evidence across turns. Its unit of measurement is therefore not surface fluency. It is the correctness of a verifiable analytical trajectory: whether the agent reads the data correctly, selects appropriate analyses, carries evidence across turns, chains dependent computations, adapts when assumptions fail, calibrates uncertainty, and grounds conclusions in the visible data and the task domain.

We build both TimeSage-MT and TimeSage because they answer different questions. TimeSage-MT defines the evaluation problem: 240 tasks, 2,680 dialogue turns, 4 levels, 8 domains, reference code, numerical gold, reasoning annotations, and outcome/capability scoring. TimeSage is a reference structured agent system for the same setting. It demonstrates one engineering pattern for combining planning, skills, executable analysis, deterministic validation, and report generation around the benchmark interface. The benchmark does not require future systems to use the TimeSage architecture. Rather, TimeSage is included to make the system-design question concrete: when does a structured skill-and-harness workflow help, and when do simpler direct or code-enabled agents suffice?

The scope of validity follows from this design. Within the benchmark contract, TimeSage-MT supports rigorous comparison of agentic systems on verifiable, multi-turn time series reasoning. It is not a benchmark for prompt aesthetics, conversational charm, marketing copy, or open-ended creativity, and we advise against interpreting it that way. It also should not be read as a deployment certificate for high-impact settings. A high leaderboard score means that an agent performed well on the released benchmark distribution; it does not remove the need for local validation, human oversight, privacy review, operational fail-safes, or domain-specific safety approval in finance, healthcare, energy, climate, manufacturing, transportation, telecom, or retail.

The aggregate outcome and capability scores should be treated as summaries, not as the sole object of research. TimeSage-MT is designed so that researchers can move from a leaderboard number to the underlying evidence: per-task scores, per-turn scores, outcome subtracks, capability subchecks, code outputs, tool traces, skill calls, reasoning-graph alignment, and reviewer/audit metadata. The intended workflow is therefore diagnostic. Rather than only chasing a higher benchmark accuracy, researchers should inspect where an agent failed: whether it misread the data, selected the wrong method, configured a parameter incorrectly, forgot an earlier result, broke a dependency chain, accepted a misleading user hint, understated uncertainty, or made an unsupported domain decision. This breakdown is what makes the benchmark useful for designing better time series agentic systems: it turns a scalar score into a map of which component of the analytical workflow needs better modeling, tooling, memory, validation, or orchestration.

\section{Limitations and Open Challenges}
\label{appendix_h}
TimeSage-MT provides a structured evaluation framework for multi-turn time series reasoning, but it does not cover the full diversity of real analytical settings. The benchmark uses bounded, scripted conversations, and therefore only partially captures long-horizon memory, collaborative analysis, evolving user goals, and cases where earlier assumptions must be revised after new evidence appears. Its domain coverage is also constrained by the availability of public, license-compatible time series data. Although TimeSage-MT spans eight real-world domains, it cannot fully represent the regulatory constraints, institutional policies, rare events, proprietary data, and causal assumptions that often shape high-stakes decisions.

Several open challenges remain for both benchmark design and agent development. Current agents still struggle with uncertainty calibration, domain grounding, and robust generalization to unfamiliar data distributions, tool interfaces, and task structures. At the same time, evaluation design remains difficult for open-ended, non-linear reasoning paths, especially when correctness depends on risk thresholds, asymmetric costs, or domain-specific judgment. Future work should expand benchmark coverage, strengthen calibration and domain-specific evaluation, compare LLM-based judging against expert review, and study whether agent performance generalizes beyond the benchmark harness.

\section{Broader Impact}
\label{appendix_i}
TimeSage-MT is intended to promote more rigorous and transparent evaluation of time series reasoning agents before they are used in consequential analytical settings. By measuring numerical accuracy, factual verification, uncertainty handling, and domain-grounded decision making, the benchmark can help identify failure modes that are not captured by single-turn forecasting or classification tests. This may support the development of more auditable AI tools for domains such as finance, healthcare, energy, climate, manufacturing, telecommunications, transportation, and retail, where time series analysis often informs recurring operational decisions.

At the same time, benchmark performance should not be interpreted as sufficient evidence for safe deployment. Automated agents may still produce plausible but incorrect analyses, understate uncertainty, miss rare events, or recommend actions that ignore local constraints and real-world costs. These risks are especially important in high-impact domains where analytical outputs can affect safety, resource allocation, patient care, infrastructure, or financial decisions. Any deployed system should therefore include human oversight, independent validation on local data, audit trails, privacy protections, operational fail-safes, and domain-specific safety review.




\end{document}

%% file: table/observation_1_2.tex
\begin{table}[t]
\centering
\caption{(a): Direct (R) vs.\ Code-Enabled (C) Reasoning. (b): Code-Enabled Reasoning without (W) vs.\ with (S) skill guidance. \textit{Top}: outcome scores across five dimensions (NA: numerical accuracy, FV: factual verification, AQ: analytical quality, DM: decision making, CC: code correctness). \textit{Bottom}: outcome scores per tier (L1--L4) and overall. \textbf{Bold} denotes the higher score in each comparison.}
\label{tab:obs_combined}

\vspace{5pt}

\setlength{\tabcolsep}{3pt}
\renewcommand{\arraystretch}{1.05}
\scriptsize
\begin{minipage}[t]{0.5\linewidth}
\centering
\textbf{(a) Direct vs.\ Code-Enabled Reasoning} \\[2pt]
\begin{tabular}{lcccccccccc}
\toprule
Model & \multicolumn{2}{c}{NA} & \multicolumn{2}{c}{FV} & \multicolumn{2}{c}{AQ} & \multicolumn{2}{c}{DM} & \multicolumn{2}{c}{CC} \\
\cmidrule(lr){2-3} \cmidrule(lr){4-5} \cmidrule(lr){6-7} \cmidrule(lr){8-9} \cmidrule(lr){10-11}
 & R & C & R & C & R & C & R & C & R & C \\
\midrule
gpt-5.3-codex & 11 & \textbf{18} & 12 & \textbf{20} & 32 & \textbf{35} & 48 & \textbf{63} & 00 & \textbf{79} \\
qwen3.5-122b  & 15 & \textbf{16} & \textbf{27} & 21 & 25 & \textbf{30} & 40 & \textbf{63} & 00 & \textbf{74} \\
\bottomrule
\end{tabular}

\medskip

\begin{tabular}{lcccccccccc}
\toprule
Model & \multicolumn{2}{c}{L1} & \multicolumn{2}{c}{L2} & \multicolumn{2}{c}{L3} & \multicolumn{2}{c}{L4} & \multicolumn{2}{c}{Overall} \\
\cmidrule(lr){2-3} \cmidrule(lr){4-5} \cmidrule(lr){6-7} \cmidrule(lr){8-9} \cmidrule(lr){10-11}
 & R & C & R & C & R & C & R & C & R & C \\
\midrule
gpt-5.3-codex & 21 & \textbf{48} & 07 & \textbf{34} & 14 & \textbf{37} & 19 & \textbf{39} & 15 & \textbf{39} \\
qwen3.5-122b  & 26 & \textbf{47} & 12 & \textbf{32} & 14 & \textbf{34} & 18 & \textbf{36} & 18 & \textbf{37} \\
\bottomrule
\end{tabular}
\label{tab:ob_1}
\end{minipage}%
\begin{minipage}[t]{0.5\linewidth}
\centering
\textbf{(b) Without vs.\ With Skill Guidance} \\[2pt]
\begin{tabular}{lcccccccccc}
\toprule
Model & \multicolumn{2}{c}{NA} & \multicolumn{2}{c}{FV} & \multicolumn{2}{c}{AQ} & \multicolumn{2}{c}{DM} & \multicolumn{2}{c}{CC} \\
\cmidrule(lr){2-3} \cmidrule(lr){4-5} \cmidrule(lr){6-7} \cmidrule(lr){8-9} \cmidrule(lr){10-11}
 & W & S & W & S & W & S & W & S & W & S \\
\midrule
gpt-5.3-codex & \textbf{18} & 17 & \textbf{20} & 19 & 35 & \textbf{61} & 63 & \textbf{76} & \textbf{79} & 77 \\
qwen3.5-122b  & \textbf{16} & \textbf{16} & \textbf{21} & 21 & \textbf{30} & \textbf{30} & 63 & \textbf{63} & 74 & \textbf{76} \\
\bottomrule
\end{tabular}

\medskip

\begin{tabular}{lcccccccccc}
\toprule
Model & \multicolumn{2}{c}{L1} & \multicolumn{2}{c}{L2} & \multicolumn{2}{c}{L3} & \multicolumn{2}{c}{L4} & \multicolumn{2}{c}{Overall} \\
\cmidrule(lr){2-3} \cmidrule(lr){4-5} \cmidrule(lr){6-7} \cmidrule(lr){8-9} \cmidrule(lr){10-11}
 & W & S & W & S & W & S & W & S & W & S \\
\midrule
gpt-5.3-codex & 48 & \textbf{54} & 34 & \textbf{37} & 37 & \textbf{44} & 39 & \textbf{45} & 39 & \textbf{45} \\
qwen3.5-122b  & 47 & \textbf{47} & 32 & \textbf{33} & 34 & \textbf{34} & 36 & \textbf{37} & 37 & \textbf{38} \\
\bottomrule
\end{tabular}
\label{tab:ob_2}
\end{minipage}

\label{tab:ob_1_2}
\end{table}

%% file: table/appendix_table/appendix_a/1_capability.tex
\begin{table}[htbp]
    \centering  
    \small
    \renewcommand\arraystretch{1.15}
    \caption{
        The nine TimeSage-MT capabilities. \textit{Family} groups capabilities into three analyst-workflow phases: \textit{Basic} (read, pick, configure a method), \textit{Reasoning} (use prior turns, chain analyses, recover from errors), and \textit{Action} (commit to a decision, calibrate confidence, ground in domain knowledge). The \textit{Level} column gives the level at which each capability becomes scorable.
    }
    \vspace{3mm}
    \label{tab:capability-taxonomy}
    \begin{tabular}{>{\centering\arraybackslash}m{0.03\textwidth} >{\centering\arraybackslash}m{0.09\textwidth} | >{\centering\arraybackslash}m{0.09\textwidth} | m{0.58\textwidth} | >{\centering\arraybackslash}m{0.05\textwidth}}
    \toprule\toprule
    \textbf{ID} & \textbf{Capability} & \textbf{Family} & \multicolumn{1}{c|}{\textbf{Operational definition}} & \textbf{Level} \\
    \hline
    C1 & Read       &                                  & Identify data properties (frequency, length, stationarity, seasonality, missingness, outliers, trend) without inventing values. & L1+ \\
    C2 & Select     & \multirow{-1}{*}{\textit{Basic}} & Pick appropriate skills from the 226-skill library given the data profile and task. & L1+ \\
    C3 & Configure  &                                  & Set skill parameters (forecast horizon, seasonal period, anomaly threshold, train/test split, etc.) and select skills. & L1+ \\ \hline
    C4 & Remember   &                                      & Use values, decisions, or artefacts from prior turns rather than recomputing or contradicting them. & L2+ \\
    C5 & Chain      & \multirow{-1}{*}{\textit{Reasoning}} & Order analyses in terms of dependency edges in the gold reasoning graph defined. & L2+ \\
    C6 & Adapt      &                                      & Recover when the user redirects, presents new evidence, or pushes a misleading instruction. & L1+ \\ \hline
    C7 & Decide     &                                   & Convert analysis into a coherent, actionable decision consistent with the analysis trace. & L3+ \\
    C8 & Calibrate  & \multirow{-1}{*}{\textit{Action}} & Report uncertainty (intervals, confidence levels, hedging language) within nominal coverage. & L3+ \\
    C9 & Ground     &                                   & Apply domain-specific knowledge (industry conventions, regulatory thresholds) when the task carries a domain. & L3+ \\
    \bottomrule\bottomrule
    \end{tabular}
\end{table}
\vspace{1em}

%% file: table/appendix_table/appendix_a/2_task.tex
\begin{table}[htbp]
    \centering  
    \small
    \renewcommand\arraystretch{1.15}
    \caption{Task-level design. Turn ranges are sampled by length bucket; the released corpus is balanced over levels and user styles.}
    \vspace{3mm}
    \label{tab:task-levels}
    \begin{tabular}{>{\centering\arraybackslash}m{0.06\textwidth} | m{0.18\textwidth} | m{0.18\textwidth} | m{0.19\textwidth} | m{0.25\textwidth}}
    \toprule\toprule
    \textbf{Level} & \multicolumn{1}{c|}{\textbf{Task role}} & \multicolumn{1}{c|}{\textbf{Dialogue length}} & \multicolumn{1}{c|}{\textbf{Reasoning graph}} & \multicolumn{1}{c}{\textbf{Task \& capabilities}} \\
    \hline
    L1 & Open exploration or single-step profile query. & 3--10 turns (short: 3--4, medium: 5--7, long: 8--10). & No in-depth analytical step required. & Profile or data-summary answer. Tests C1--C3 and C6 when a correction-style turn is present. \\ \hline
    L2 & Multi-skill analysis over a visible time-series slice. & 5--12 turns (short: 5--7, medium: 8--10, long: 11--12). & 2--4 analytical steps with at least one dependency. & Per-turn analytical outputs and at least one cross-turn finding. Tests C1--C6. \\ \hline
    L3 & Grounded synthesis across methods or evidence types. & 8--15 turns (short: 8--10, medium: 11--13, long: 14--15). & 4--6 analytical steps with domain sepecific dependencies. & Structured synthesis gold plus turn-level findings. Tests C1--C9 except final decision-specific structure. \\ \hline
    L4 & Full decision workflow. & 12--20 turns (short: 12--14, medium: 15--17, long: 18--20). & 6--10 analytical steps ending in a decision node. & Decision with recommendation, rationale, confidence, and risks. Tests all C1--C9. \\
    \bottomrule\bottomrule
    \end{tabular}
\end{table}
\vspace{1em}

%% file: table/appendix_table/appendix_b/pipeline.tex
\begin{table}[htbp]
    \centering  
    \small
    \renewcommand\arraystretch{1.15}
    \caption{Data Generation pipeline phase overview. ``LLM?'' indicates whether the phase calls a language model. ``Det.?'' indicates whether the phase performs deterministic checks. ``Human?'' indicates whether the phase involves human reviewers. P1 is the generation pipeline (5 sub-stages, S1--S5); P2--P5 are the quality-control cycle.}
    \vspace{3mm}
    \label{tab:app_pipeline_overview}
    \begin{tabular}{>{\centering\arraybackslash}m{0.05\textwidth} | m{0.23\textwidth} | >{\centering\arraybackslash}m{0.06\textwidth} | >{\centering\arraybackslash}m{0.06\textwidth} | >{\centering\arraybackslash}m{0.075\textwidth} | m{0.34\textwidth}}
    \toprule\toprule
    \textbf{Phase} & \multicolumn{1}{c|}{\textbf{Sub-stage}} & \textbf{LLM?} & \textbf{Det.?} & \textbf{Human?} & \multicolumn{1}{c}{\textbf{Purpose}} \\
    \hline
     & S1 Data Selection        & ---        & \ding{51} & ---        & Curate 65 whitelisted real sources \\
     & S2 Profiling             & ---        & \ding{51} & ---        & 11-metric data profile + 80/20 split \\
    \multirow{-1}{*}{P1} & S3 Level \& Style Assign & ---        & \ding{51} & ---        & Quota-balanced assignment \\
     & S4 Reasoning Path        & ---        & \ding{51} & ---        & Build typed DAG per task \\
     & S5 Dialogue Generation   & \ding{51}  & \ding{51} & ---        & LLM dialogue, rule-based post-checks \\ \hline
    P2 & Reproducibility audit   & ---       & \ding{51} & ---        & 46 deterministic checks across 10 categories \\ \hline
    P3 & LLM cross-family review & \ding{51} & ---       & ---        & 4 critical-checklist items, anti-bias model \\ \hline
    P4 & Human review            & ---       & ---       & \ding{51}  & 7-item checklist via dashboard \\ \hline
    P5 & Final check             & ---       & \ding{51} & \ding{51}  & Re-run audits + sandbox + human spot-check \\
    \bottomrule\bottomrule
    \end{tabular}
\end{table}
\vspace{1em}

%% file: table/appendix_table/appendix_b/P1/1_whitelist_checklist.tex
\begin{table}[htbp]
    \centering  
    \small
    \renewcommand\arraystretch{1.15}
    \caption{P1.S1 Data selection checklist.}
    \vspace{3mm}
    \label{tab:whitelist_check}
    \begin{tabular}{m{0.20\textwidth} | m{0.60\textwidth} | >{\centering\arraybackslash}m{0.10\textwidth}}
    \toprule\toprule
    \multicolumn{1}{c|}{\textbf{Checklist item}} & \multicolumn{1}{c|}{\textbf{Subchecks}} & \textbf{Uses LLM} \\
    \hline
    Retrieval and identity & Source ID, canonical URL or local path, retrieval timestamp, SHA-256 fingerprint, repository-relative path. & No \\ \hline
    License and policy & License status is permitted, source is not marked unusable, contamination-risk flag is present. & No \\ \hline
    Parseability & File parses as tabular data; datetime-like column exists; at least one numeric signal exists. & No \\ \hline
    Level feasibility & Minimum row count by level: L1 at least 50, L2 at least 100, L3 at least 500, L4 at least 1000. & No \\ \hline
    Manifest output & Emits a source manifest that will be used by S2--S5 and for later stage P2/P5 checks. & No \\
    \bottomrule\bottomrule
    \end{tabular}
\end{table}
\vspace{1em}

%% file: table/appendix_table/appendix_b/P1/2_profile_checklist.tex
\begin{table}[htbp]
    \centering
    \small
    \renewcommand\arraystretch{1}
    \setlength{\extrarowheight}{2pt}
    \caption{P1.S2 deterministic time series profile checklist. All fields are computed without LLM calls and are used to decide whether a source supports a given level, task family, and scoring protocol.}
    \vspace{3mm}
    \label{tab:profile_checklist}
    \begin{tabular}{m{0.13\textwidth} | m{0.34\textwidth} | m{0.32\textwidth} | >{\centering\arraybackslash}m{0.10\textwidth}}
    \toprule\toprule
    \multicolumn{1}{c|}{\textbf{Profile field}} & \multicolumn{1}{c|}{\textbf{Check}} & \multicolumn{1}{c|}{\textbf{Downstream use}} & \textbf{Uses LLM} \\
    \hline
    Time axis and frequency & Detects timestamp column, ordering, duplicate timestamps, irregular gaps, and inferred sampling frequency. & Validates source eligibility and determines resampling, forecasting horizon, and analysis window size. & No \\ \hline
    Length and split capacity & Counts usable observations before and after the 80/20 visible/held-out split; records visible end, held-out start, and total length. & Ensures short, medium, and long tasks have enough visible context and scorer-only held-out gold. & No \\ \hline
    Dimensionality & Counts numeric signals, target candidates, covariates, and label columns when present. & Selects feasible task families such as forecasting, classification, regression, anomaly detection, and causal analysis. & No \\ \hline
    Missingness & Computes missing-value rate, missing runs, and column-level missingness. & Determines whether imputation, data-quality, or preprocessing tasks are feasible. & No \\ \hline
    Outliers and noise & Estimates robust outlier rate and noise-to-signal ratio using deterministic statistics. & Supports anomaly, robustness, preprocessing, and diagnostic tasks. & No \\ \hline
    Autocorrelation & Computes lag-1 autocorrelation and related temporal dependence summaries. & Determines whether forecasting, lag analysis, and temporal-dependence tasks are meaningful. & No \\ \hline
    Trend and seasonality & Estimates trend strength and seasonality strength when series length and frequency permit. & Enables decomposition, seasonal forecasting, trend interpretation, and domain-grounded synthesis tasks. & No \\ \hline
    Stationarity & Runs deterministic stationarity diagnostics and summary statistics. & Supports transformation, differencing, forecasting-model choice, and profile questions. & No \\ \hline
    Label or event availability & Detects class labels, anomaly labels, change-point labels, event times, or intervention fields when present. & Enables classification, anomaly scoring, event-sequence, change-point, and causal task families. & No \\ \hline
    Feasible task families & Combines profile constraints into a source-level feasibility mask. & Prevents generation of tasks whose gold answer or metric cannot be computed reproducibly. & No \\ \hline
    Artifact provenance & Records visible CSV path, visible CSV SHA-256, visible/held-out split metadata, and profile JSON. & Supports deterministic replay, leakage checks, and later P2/P5 reproducibility audits. & No \\
    \bottomrule\bottomrule
    \end{tabular}
\end{table}
\vspace{1em}

%% file: table/appendix_table/appendix_b/P1/3_task_assign.tex
\begin{table}[htbp]
    \centering  
    \small
    \renewcommand\arraystretch{1.4}
    \setlength{\extrarowheight}{2pt}
    \caption{P1.S3 Task assignment checklist.}
    \vspace{3mm}
    \label{tab:task_assign_checklist}
    \begin{tabular}{m{0.20\textwidth} | m{0.62\textwidth} | >{\centering\arraybackslash}m{0.10\textwidth}}
    \toprule\toprule
    \multicolumn{1}{c|}{\textbf{Checklist item}} & \multicolumn{1}{c|}{\textbf{Subchecks}} & \textbf{Uses LLM} \\
    \hline
    Level quota & Exactly 60 tasks per level, for 240 total tasks. & No \\ \hline
    Length quota & Exactly 20 short, 20 medium, and 20 long assignments per level. & No \\ \hline
    Style quota & Per level: 30 open, 20 guided-correct, 10 guided-wrong. & No \\ \hline
    Domain quota & L3 and L4 assignments are distributed over finance, healthcare, energy, climate, retail, manufacturing, transportation, and telecom when supported by the source. & No \\ \hline
    Feasibility check & Assignment family must be feasible for the selected source profile and split length. & No \\
    \bottomrule\bottomrule
    \end{tabular}
\end{table}
\vspace{1em}

%% file: table/appendix_table/appendix_b/P1/4_reasoning_graph.tex
\begin{table}[htbp]
    \centering  
    \small
    \renewcommand\arraystretch{1.4}
    \setlength{\extrarowheight}{2pt}
    \caption{P1.S4 Reasoning Path Construction checklist.}
    \vspace{3mm}
    \label{tab:reasoning_path_checklist}
    \begin{tabular}{m{0.22\textwidth} | m{0.62\textwidth} | >{\centering\arraybackslash}m{0.10\textwidth}}
    \toprule\toprule
    \multicolumn{1}{c|}{\textbf{Checklist group}} & \multicolumn{1}{c|}{\textbf{Subchecks}} & \textbf{Uses LLM} \\
    \hline
    Structural validity & Graph is acyclic; node IDs are unique; every edge endpoint exists; graph is reachable; topological depth and width are within level limits. & No \\ \hline
    Level shape & L1 has no required graph; L2 has 2--4 analytical steps; L3 has 4--6 steps and a synthesis path; L4 has 6--10 steps and a final decision path. & No \\ \hline
    Skill validity & Required and allowed skills are present in the released skill registry; node capability tags are valid; output type is supported by the source profile. & No \\ \hline
    Dependency semantics & Edge types are valid; edge rationale is populated; required and optional dependencies are balanced; graph weights are normalized. & No \\ \hline
    Capability coverage & Capability tags satisfy level-specific requirements, including C6 gating for guided-wrong assignments and C7--C9 for synthesis or decision tasks. & No \\
    \bottomrule\bottomrule
    \end{tabular}
\end{table}
\vspace{1em}

%% file: table/appendix_table/appendix_b/P1/5_data_generation.tex
\begin{table}[htbp]
    \centering  
    \small
    \renewcommand\arraystretch{1.4}
    \setlength{\extrarowheight}{2pt}
    \caption{P1.S5 Dialogue generation checklist.}
    \vspace{3mm}
    \label{tab:dialogue_generation_checklist}
    \begin{tabular}{m{0.18\textwidth} | m{0.56\textwidth} | m{0.20\textwidth}}
    \toprule\toprule
    \multicolumn{1}{c|}{\textbf{Checklist group}} & \multicolumn{1}{c|}{\textbf{Subchecks}} & \multicolumn{1}{c}{\textbf{Mechanism}} \\
    \hline
    Dialogue schema & Strict turn IDs, first turn is user, user/agent alternation, turn count within level range, valid JSON schema. & Deterministic \\ \hline
    Graph coverage & All graph nodes are covered; analytical agent turns map to valid nodes; expected skills appear in agent turns. & Deterministic \\ \hline
    Skill and parameter fields & Canonical skill names, expected skills, skills invoked, expected parameters, output type, and reference-code sentinel. & LLM authored, deterministic validation \\ \hline
    Reference code & 5--25 line Python snippets for computational turns, visible CSV only, no held-out-row references, printed numeric claims captured in stdout. & LLM authored, deterministic execution/validation \\ \hline
    Cross-turn structure & L2+ has valid prior-turn references; L3 has synthesis; L4 has final decision JSON; guided-wrong tasks have an adversarial turn and agent correction. & LLM authored, deterministic validation \\ \hline
    Gold extraction & Forecast values, anomaly lists, change points, classification labels, regression targets, imputation cells, causal edges, counterfactuals, synthesis findings, and confidence intervals are produced by extractors rather than copied from the dialogue. & Deterministic \\ \hline
    Leakage guard & No user-visible text names held-out row counts, split boundary integers, total length, or held-out values. & Deterministic \\
    \bottomrule\bottomrule
    \end{tabular}
\end{table}
\vspace{1em}

%% file: table/appendix_table/appendix_b/P2/1_reproduce_checklist.tex
\begin{table}[htbp]
    \centering  
    \small
    \renewcommand\arraystretch{1.2}
    \setlength{\extrarowheight}{2pt}
    \caption{P2 deterministic audit categories and subchecks. ``$n$'' is the number of checks in each category, summing to 46. Every check returns pass/fail/not-applicable and is reproducible.}
    \vspace{3mm}
    \label{tab:p2_checklist}
    \begin{tabular}{>{\centering\arraybackslash}m{0.03\textwidth} | m{0.21\textwidth} | >{\centering\arraybackslash}m{0.03\textwidth} | m{0.65\textwidth}}
    \toprule\toprule
    \textbf{\#} & \multicolumn{1}{c|}{\textbf{Category}} & \textbf{$n$} & \multicolumn{1}{c}{\textbf{Subchecks}} \\
    \hline
    1 & Data integrity & 5 & Source SHA format, source SHA matches disk, permitted license, contamination flag present. \\ \hline
    2 & Split integrity & 4 & 80/20 split math, visible CSV exists, visible CSV row count, visible CSV SHA, imputation CSV artifact, no held-out rows in visible CSV. \\ \hline
    3 & Assignment validity & 3 & Level minimum rows, turn count in range, step count in range, domain rule, turn sequence invariants. \\ \hline
    4 & Graph validity & 5 & DAG validity, unique node IDs, valid edge endpoints, valid skill references, turn-to-node mappings, level graph shape. \\ \hline
    5 & Reference code & 6 & Python parses, skill sentinel present, analytical skills in library, no held-out-row references, language tag, executable flag. \\ \hline
    6 & Gold provenance & 5 & Gold provenance fields, non-null gold for typed targets, held-out row indices in range, non-empty indices, forecast horizon match, expected parameters populated, key finding on synthesis/decision turns, verification spec present when key finding is present. \\ \hline
    7 & Narrative grounding & 5 & Narrative is grounded in reference stdout, analytical text is non-empty, no visibility leakage in dialogue. \\ \hline
    8 & Metadata consistency & 4 & Required metadata fields, quality gate decision, quality gate pass, attempts in accepted range. \\ \hline
    9 & Cross-task consistency & 5 & Task ID uniqueness and source ID present in manifest. \\ \hline
    10 & Knowledge integrity & 4 & Knowledge-pack path exists and knowledge-pack SHA is recorded. \\ \hline
     & \multicolumn{1}{c|}{\textbf{Total}} & \textbf{46} & \\
    \bottomrule\bottomrule
    \end{tabular}
\end{table}
\vspace{1em}

%% file: table/appendix_table/appendix_b/P4/1_human_review_checklist.tex
\begin{table}[htbp]
    \centering  
    \small
    \renewcommand\arraystretch{1.2}
    \setlength{\extrarowheight}{2pt}
    \caption{P4 human review checklist.}
    \label{tab:human_review_checklist}
    \vspace{3mm}
    \begin{tabular}{m{0.18\textwidth} | m{0.56\textwidth} | m{0.18\textwidth}}
    \toprule\toprule
    \multicolumn{1}{c|}{\textbf{Review area}} & \multicolumn{1}{c|}{\textbf{Human checklist}} & \multicolumn{1}{c}{\textbf{Mechanism}} \\
    \hline
    Task context & Source, domain, level, style, turn count, visible/held-out split metadata, and profile look coherent. & Human \\ \hline
    Dialogue quality & User turns are natural and open-loop evaluable; agent turns are grounded, non-leaky, and cover the graph. & Human \\ \hline
    Analytical validity & Reference code, printed numbers, key findings, and gold targets match the intended analytical step. & Human + deterministic reports \\ \hline
    Capability coverage & C1--C9 tags and expected skills/parameters match the level and reasoning path. & Human \\ \hline
    Revision routing & Reviewer can edit JSON, request S5 regeneration, reject, or approve. & Human + dashboard actions \\
    \bottomrule\bottomrule
    \end{tabular}
\end{table}
\vspace{1em}

%% file: table/appendix_table/appendix_b/P5/1_final_check.tex
\begin{table}[htbp]
    \centering  
    \small
    \renewcommand\arraystretch{1.4}
    \setlength{\extrarowheight}{2pt}
    \caption{P5 final-release checklist.}
    \vspace{3mm}
    \label{tab:final_release_checklist}
    \begin{tabular}{m{0.22\textwidth} | m{0.55\textwidth} | m{0.20\textwidth}}
    \toprule\toprule
    \multicolumn{1}{c|}{\textbf{Gate}} & \multicolumn{1}{c|}{\textbf{Subchecks}} & \multicolumn{1}{c}{\textbf{Mechanism}} \\
    \hline
    Human approval state & Task has an assigned P4 approval and is not in revision or rejected status. & Human state checked deterministically \\ \hline
    P2 rerun & Data, visible/held-out split, assignment, graph, code, gold, narrative, metadata, cross-task, and knowledge-pack checks rerun over final files. & Deterministic \\ \hline
    P3 rerun or import & If enabled, cross-family LLM review is rerun; otherwise previous review artifacts are imported and failures remain blocking. & Optional LLM \\ \hline
    Reference-code execution & Sandbox re-executes snippets when configured and records stdout/stderr, parse errors, and execution failures. & Deterministic sandbox \\ \hline
    Routing & Passing tasks copied to final passed corpus; failures copied to quarantine; rerun queues created for regenerable failures. & Deterministic \\ \hline
    Manifest & Final manifest records task IDs, SHAs, audit status, reviewer state, and release location. & Deterministic \\
    \bottomrule\bottomrule
    \end{tabular}
\end{table}
\vspace{1em}

%% file: table/appendix_table/appendix_c/1_outcome_evaluation.tex
\begin{table}[htbp]
    \centering  
    \small
    \renewcommand\arraystretch{1.15}
    \caption{Outcome metrics, level applicability, and scoring rules. MASE denotes mean absolute scaled error; F1 is harmonic mean of precision and recall; 1-NRMSE is one minus normalized root-mean-square error; CI denotes confidence interval; ARI denotes adjusted Rand index.}
    \vspace{3mm}
    \label{tab:outcome_metrics}
    \begin{tabular}{m{0.18\textwidth} | m{0.30\textwidth} | m{0.32\textwidth} | >{\centering\arraybackslash}m{0.10\textwidth}}
    \toprule\toprule
    \multicolumn{1}{c|}{\textbf{Metric}} & \multicolumn{1}{c|}{\textbf{Calculation}} & \multicolumn{1}{c|}{\textbf{Subchecks and target types}} & \textbf{Uses LLM} \\
    \hline
    Numerical accuracy & Typed scorer compares extracted agent output to expected output value. Scores are normalized to $[0,1]$. & Forecast MASE threshold; anomaly F1; change-point F1 with tolerance; classification accuracy; regression/imputation 1-NRMSE; causal edge F1; counterfactual point/CI match; generation or editing statistic preservation; clustering ARI; profile attribute match. & No \\ \hline
    Factual verification & Measures whether data-grounded key facts from reference stdout, gold fields, and key findings appear in the response. & Numeric overlap within 5\% relative tolerance; key-finding coverage; optional local embedding similarity when sentence-transformers is installed. & No \\ \hline
    Analytical quality & LLM judge scores whether the analysis is correct, insightful, and supported by the actual data, using reference stdout and one valid answer for calibration. & Factual correctness, analytical depth, conclusion soundness; arithmetic mean of the three judge axes. & Yes \\ \hline
    Code correctness & Executes agent code when present and compares behavior to reference execution. & $0.5 \times$ executes-cleanly $+ 0.5 \times$ stdout number match, where stdout match is the fraction of reference numbers recovered within tolerance. & No \\ \hline
    Decision making & L4-only LLM judge scores the final decision against the actual data and reference stdout, not against a single mandatory label. & Decision validity, reasoning quality, completeness; arithmetic mean of the three judge axes. & Yes \\
    \bottomrule\bottomrule
    \end{tabular}
\end{table}
\vspace{1em}

%% file: table/appendix_table/appendix_c/2_capability_evaluation.tex
\begin{table}[htbp]
    \centering  
    \small
    \renewcommand\arraystretch{1.2}
    \caption{Capability subchecks and aggregation.}
    \vspace{3mm}
    \label{tab:capability_subchecks}
    \begin{tabular}{>{\centering\arraybackslash}m{0.04\textwidth} | m{0.30\textwidth} | m{0.50\textwidth} | >{\centering\arraybackslash}m{0.10\textwidth}}
    \toprule\toprule
    \textbf{ID} & \multicolumn{1}{c|}{\textbf{Subchecks}} & \multicolumn{1}{c|}{\textbf{Calculation details}} & \textbf{Uses LLM} \\
    \hline
    C1 & Profile/data read checks & Matches frequency, stationarity, seasonality, missing rate, trend, outlier count, and summary fields recovered from structured output, narrative, stdout, or cross-turn memory. & No \\ \hline
    C2 & Required-skill recall and alternative acceptance & Compares expected skills with invoked skills recovered from traces, code AST, narrative, or stdout. Allows aliases, family equivalence, and local embedding match. & No \\ \hline
    C3 & Exact or tolerance parameter checks & Compares expected parameters such as horizon, lags, target column, threshold, visible/held-out split, and tolerance to recovered parameters. & No \\ \hline
    C4 & Reference, no-redundancy, combined memory & Checks whether the agent references prerequisite turns and avoids needless repeated analysis. & No \\ \hline
    C5 & DAG (Directed Acyclic Graph) order and logical dependency & Penalizes dependency-order violations; code-interpreter outputs also receive numeric-overlap dependency checks. & No \\ \hline
    C6 & Correction behavior & Scores whether the agent changes skill, parameter, or narrative when the user supplies a misleading or corrective turn. & No \\ \hline
    C7 & Citation rate, required information coverage, structure completeness, cross-turn coherence, logical coherence & Rule checks cover citations and required fields; the LLM coherence judge checks whether the conclusion is supported by the analysis. & Hybrid \\ \hline
    C8 & CI coverage, CI width, uncertainty expression & Checks whether reported uncertainty includes the gold value, is not excessively wide, and is expressed in language or structured fields. & No \\ \hline
    C9 & Domain term rate, threshold correctness, domain reasoning quality & Rule checks cover domain vocabulary and thresholds; LLM judge checks whether the domain interpretation is sound. & Hybrid \\
    \bottomrule\bottomrule
    \end{tabular}
\end{table}
\vspace{1em}

%% file: table/appendix_table/appendix_c/3_joint.tex
\begin{table}[htbp]
    \centering  
    \small
    \renewcommand\arraystretch{1.2}
    \caption{Interpretation of the joint outcome--capability evaluation axes.}
    \vspace{3mm}
    \label{tab:joint_outcome_capability}
    \begin{tabular}{>{\centering\arraybackslash}m{0.15\textwidth} | >{\centering\arraybackslash}m{0.16\textwidth} | m{0.58\textwidth}}
    \toprule\toprule
    \textbf{Outcome score} & \textbf{Capability score} & \multicolumn{1}{c}{\textbf{Interpretation}} \\
    \hline
    High & High & The agent reaches the correct answer and follows the expected analytical workflow. This is the desired behavior for reliable deployment. \\ \hline
    High & Low & The final answer is correct, but the trace lacks required skills, dependency tracking, grounding, or correction behavior. This may indicate shortcutting, brittle reasoning, or lucky agreement with the gold answer. \\ \hline
    Low & High & The agent follows many intended analytical steps but fails to produce the correct final output. This reveals execution, parameter, numerical, or synthesis failures despite reasonable process behavior. \\ \hline
    Low & Low & The agent fails both the final-answer target and the expected analytical behavior. This indicates a broad mismatch with the benchmark task. \\
    \bottomrule\bottomrule
    \end{tabular}
\end{table}
\vspace{1em}

%% file: table/appendix_table/appendix_d/1_comparison_agentic_system.tex
\begin{table}[htbp]
    \centering  
    \small
    \renewcommand\arraystretch{1.2}
    \setlength{\extrarowheight}{2pt}
    \caption{Comparison of evaluated agentic systems.}
    \vspace{3mm}
    \label{tab:agentic_systems}
    \begin{tabular}{m{0.16\textwidth} | m{0.24\textwidth} | m{0.20\textwidth} | m{0.28\textwidth}}
    \toprule\toprule
    \multicolumn{1}{c|}{\textbf{System}} & \multicolumn{1}{c|}{\textbf{Tool access}} & \multicolumn{1}{c|}{\textbf{State across turns}} & \multicolumn{1}{c}{\textbf{Purpose in benchmark}} \\
    \hline
    Direct Answering & 
    None & 
    Conversation text only & 
    Tests prompt-only reasoning and memorized analytical priors. \\ \hline
    Code-Enabled Reasoning & 
    Python sandbox & 
    Prior user turns and generated outputs & 
    Tests whether executable code improves numerical accuracy and grounding. \\ \hline
    Skill-Guided Code Reasoning & 
    \vspace{2pt}
    \begin{itemize}[leftmargin=*,nosep,topsep=0pt]
        \item Python sandbox
        \item 226-skill library
    \end{itemize}
    \vspace{2pt} & 
    Prior user turns, generated outputs, and skill calls & 
    Tests the marginal value of a curated time series skill. \\ \hline
    TimeSage & 
    \vspace{2pt}
    \begin{itemize}[leftmargin=*,nosep,topsep=0pt]
        \item Skills and tools
        \item MCP adapters
        \item Validation/profiling
        \item Planner, Executor, Evaluator, Reporter
    \end{itemize}
    \vspace{2pt} & 
    Structured multi-turn session state & 
    Tests the complete agentic system under the same open-loop benchmark protocol. \textbf{Question}: do orchestration + skills compound the gains?\\
    \bottomrule\bottomrule
    \end{tabular}
\end{table}

%% file: table/appendix_table/appendix_e/1_skill_library.tex
\begin{table}[htbp]
    \centering
    \scriptsize
    \renewcommand\arraystretch{1.2}
    \setlength{\extrarowheight}{2pt}
    \caption{Skill-library inventory by category.}
    \vspace{3mm}
    \label{tab:skill_inventory}
    \begin{tabular}{>{\centering\arraybackslash}m{0.20\columnwidth} | >{\centering\arraybackslash}m{0.06\columnwidth} | >{\centering\arraybackslash}m{0.20\columnwidth} | >{\centering\arraybackslash}m{0.06\columnwidth} | >{\centering\arraybackslash}m{0.20\columnwidth} | >{\centering\arraybackslash}m{0.06\columnwidth}}
    \toprule\toprule
    \textbf{Category} & \textbf{Count} & \textbf{Category} & \textbf{Count} & \textbf{Category} & \textbf{Count} \\
    \hline
    anomaly detection & 15 & bayesian & 3 & causal and change-point detection & 6 \\ \hline
    classification & 13 & clustering & 3 & cointegration & 3 \\ \hline
    correlation & 4 & decomposition & 5 & distances & 5 \\ \hline
    exploratory data analysis & 7 & evaluation & 6 & feature engineering & 2 \\ \hline
    deep-learning forecasting & 33 & foundation-model forecasting & 10 & machine-learning forecasting & 4 \\ \hline
    forecasting services & 9 & statistical forecasting & 8 & hierarchical & 5 \\ \hline
    imputation & 15 & intermittent demand & 6 & interpretability & 3 \\ \hline
    long memory & 3 & motif discovery & 5 & numerical & 8 \\ \hline
    pattern detection & 5 & preprocessing & 7 & retrieval & 1 \\ \hline
    sandbox & 1 & segmentation & 3 & similarity search & 3 \\ \hline
    spectral & 3 & symbolic & 6 & synthetic & 3 \\ \hline
    time-series regression & 3 & uncertainty & 5 & volatility & 5 \\
    \bottomrule\bottomrule
    \end{tabular}
\end{table}

%% file: table/appendix_table/appendix_e/2_tool.tex
\begin{table}[htbp]
    \centering
    \small
    \renewcommand\arraystretch{1.2}
    \setlength{\extrarowheight}{2pt}
    \caption{Tool and MCP inventory.}
    \vspace{3mm}
    \label{tab:tool_mcp_inventory}
    \begin{tabular}{m{0.22\columnwidth} | m{0.70\columnwidth}}
    \toprule\toprule
    \multicolumn{1}{c|}{\textbf{Category}} & \multicolumn{1}{c}{\textbf{Components}} \\
    \hline
    Core tools & \texttt{bash}, \texttt{python\_execute}, \texttt{str\_replace\_editor}, \texttt{file\_operators}, \texttt{create\_chat\_completion}, \texttt{planning}, \texttt{ask\_human}, \texttt{terminate}. \\ \hline
    Research and browser tools & \texttt{mcp}, \texttt{crawl4ai}, \texttt{tool\_collection}, \texttt{browser\_use\_tool}, \texttt{web\_search}. \\ \hline
    Visualization and sandbox tools & \texttt{data\_visualization}, \texttt{visualization\_prepare}, \texttt{computer\_use\_tool}, \texttt{sandbox/sb\_browser\_tool}, \texttt{sandbox/sb\_files\_tool}, \texttt{sandbox/sb\_shell\_tool}, \texttt{sandbox/sb\_vision\_tool}. \\ \hline
    MCP skill adapter & Exposes all skills as schema-driven tools; supports standard input validation, CSV loading, asynchronous execution, and structured \texttt{SkillResult} conversion. \\ \hline
    MCP resources and prompts & Data-source resources and reusable prompts are registered so agents can discover benchmark assets and method documentation. \\
    \bottomrule\bottomrule
    \end{tabular}
\end{table}

%% file: table/appendix_table/appendix_e/3_harness.tex
\begin{table}[htbp]
    \centering
    \small
    \renewcommand\arraystretch{1}
    \setlength{\extrarowheight}{2pt}
    \caption{Harness components and rationale.}
    \vspace{3mm}
    \label{tab:harness_components}
    \begin{tabular}{m{0.22\textwidth} | m{0.43\textwidth} | m{0.30\textwidth}}
    \toprule\toprule
    \multicolumn{1}{c|}{\textbf{Component}} & \multicolumn{1}{c|}{\textbf{Engineering role}} & \multicolumn{1}{c}{\textbf{Why it matters}} \\
    \hline
    Open-loop dialogue renderer & Presents only user turns to evaluated agents, not scripted gold agent turns. & Prevents agents from copying the reference dialogue and makes each user turn independently evaluable. \\ \hline
    Visibility contract & Supplies visible CSVs containing only the 80\% visible slice, while the 20\% held-out rows remain scorer-only. & Prevents leakage and supports deterministic hidden gold extraction. \\ \hline
    Reference-code runner & Executes stored snippets and captures stdout/stderr, parse errors, and execution status. & Makes numerical claims auditable and reproducible. \\ \hline
    Skill adapter & Provides canonical skill calls, parameter schemas, artifacts, metrics, and diagnostics. & Gives agents reusable time series capabilities without changing the evaluation contract. \\ \hline
    Code introspector & Recovers logical skills and parameters from generated Python AST and printed outputs. & Allows fair capability scoring for agents that use code rather than explicit skill traces. \\ \hline
    Session state manager & Tracks conversation history, prior user turns, intermediate outputs, tool calls, and task metadata. & Enables multi-turn memory and chaining while preserving the benchmark's open-loop design. \\ \hline
    Planner executor evaluator reporter loop & Plans a DAG, executes skills/tools, validates outputs, retries failures, and synthesizes the answer. & Separates method selection, execution, validation, and explanation in TimeSage. \\ \hline
    Audit manifests & Store task IDs, source SHAs, visible CSV SHAs, review state, audit status, and release location. & Supports end-to-end reproducibility and corpus diffing. \\ \hline
    Dashboards & Provide P4 annotation, release monitoring, leaderboard views, and interactive agent sessions. & Makes construction and evaluation inspectable by humans. \\
    \bottomrule\bottomrule
    \end{tabular}
\end{table}

%% file: table/appendix_table/appendix_f_result/1_performance_tier.tex
\begin{table}[htbp]
\centering
\small
\renewcommand\arraystretch{1.4}
\caption{Overall performance by difficulty tier. Best result per column in \textbf{bold}; second-best \underline{underlined}.}
\vspace{3mm}
\label{tab:perf-tier}
\begin{tabular}{l | ccccc}
\toprule\toprule
\textbf{Model} & \textbf{L1} & \textbf{L2} & \textbf{L3} & \textbf{L4} & \textbf{Overall} \\
\hline
Claude-Sonnet-4.6        & \textbf{49.61} & \textbf{38.30} & \underline{35.89} & \underline{37.96} & \underline{40.44} \\
Gemini-3-Flash-Preview   & 44.08 & 31.74 & 31.58 & 33.45 & 35.21 \\
GLM-5.1                  & \underline{49.11} & \underline{37.50} & \textbf{36.50} & \textbf{38.87} & \textbf{40.50} \\
GPT-5.3-Codex            & 48.86 & 34.96 & 36.47 & 37.69 & 39.49 \\
MiniMax-M2.7             & 45.29 & 33.42 & 31.62 & 35.44 & 36.44 \\
Qwen-3.5-122B-A10B       & 47.18 & 33.63 & 32.41 & 36.13 & 37.34 \\
\bottomrule\bottomrule
\end{tabular}
\end{table}
\vspace{1em}

%% file: table/appendix_table/appendix_f_result/2_performance_outdim.tex
\begin{table}[htbp]
\centering
\small
\renewcommand\arraystretch{1.4}
\caption{Performance by output dimension. NA: numerical accuracy; FV: factual verification; AQ: analytical quality; DM: decision making; CC: code correctness.}
\vspace{3mm}
\label{tab:perf-outdim}
\begin{tabular}{l | ccccc}
\toprule\toprule
\textbf{Model} & \textbf{NA} & \textbf{FV} & \textbf{AQ} & \textbf{DM} & \textbf{CC} \\
\hline
Claude-Sonnet-4.6        & \underline{17.68} & \textbf{29.94} & 28.75 & 59.38 & \textbf{78.51} \\
Gemini-3-Flash-Preview   & 15.25 & 16.66 & 31.10 & 52.69 & 72.39 \\
GLM-5.1                  & \textbf{18.17} & \underline{28.79} & 29.48 & 62.88 & \underline{78.06} \\
GPT-5.3-Codex            & 17.70 & 21.18 & \textbf{35.14} & 59.45 & 77.79 \\
MiniMax-M2.7             & 15.54 & 22.28 & 27.87 & \textbf{64.08} & 71.74 \\
Qwen-3.5-122B-A10B       & 15.75 & 22.77 & \underline{29.89} & \underline{63.93} & 72.89 \\
\bottomrule\bottomrule
\end{tabular}
\end{table}
\vspace{1em}

%% file: table/appendix_table/appendix_f_result/3_performance_capability.tex
\begin{table}[htbp]
\centering
\scriptsize
\renewcommand\arraystretch{1.4}
\caption{Performance by capability (C1--C9).}
\vspace{3mm}
\label{tab:perf-capability}
\setlength{\tabcolsep}{4pt}
\begin{tabular}{l | ccccccccc}
\toprule\toprule
\textbf{Model} & \textbf{C1} & \textbf{C2} & \textbf{C3} & \textbf{C4} & \textbf{C5} & \textbf{C6} & \textbf{C7} & \textbf{C8} & \textbf{C9} \\
\hline
Claude-Sonnet-4.6      & 26.60 & 44.19 & \underline{19.77} & \underline{88.81} & \underline{99.54} & \underline{66.67} & 68.49 & \textbf{44.59} & \textbf{70.08} \\
Gemini-3-Flash-Prev.   & 22.18 & \underline{52.17} & 19.34 & 82.85 & 94.08 & 66.15 & 64.51 & 42.06 & 55.14 \\
GLM-5.1                & \textbf{27.49} & 48.48 & \textbf{20.05} & 84.76 & 99.47 & \textbf{69.58} & 68.40 & 42.34 & \underline{66.06} \\
GPT-5.3-Codex          & \underline{25.32} & 42.90 & 19.89 & 80.51 & 98.54 & 55.94 & \textbf{77.12} & 42.34 & 68.09 \\
MiniMax-M2.7           & 24.89 & 49.51 & 19.69 & 86.17 & \textbf{99.82} & 58.96 & 68.72 & \underline{43.90} & 60.77 \\
Qwen-3.5-122B-A10B     & 22.46 & \textbf{53.05} & 19.38 & \textbf{88.83} & 99.52 & 60.42 & \underline{68.81} & 42.62 & 55.72 \\
\bottomrule\bottomrule
\end{tabular}
\end{table}
\vspace{1em}

%% file: table/appendix_table/appendix_f_result/4_direct_vs_code.tex
%

\definecolor{deltapos}{RGB}{0,130,60}
\definecolor{deltaneg}{RGB}{200,30,30}

\newcommand{\dpos}[1]{\textcolor{deltapos}{$+$#1}}
\newcommand{\dneg}[1]{\textcolor{deltaneg}{$-$#1}}

\begin{table}[htbp]
\centering
\scriptsize
\renewcommand\arraystretch{1.3}
\caption{Direct Answering vs.\ Code-Enabled Reasoning by tier and output dimension, reported as percentages (\%). For each (model, tier) block, the three rows report \textit{Direct}, \textit{Code}, and $\Delta=$\,Code$-$Direct. \textcolor{deltapos}{\textbf{Green}} indicates Code-Enabled Reasoning improves over Direct Answering; \textcolor{deltaneg}{\textbf{red}} indicates degradation. Sample size $n{=}25$ per tier, $n{=}100$ overall. ``---'' indicates the dimension is not evaluated at that tier.}
\vspace{3mm}
\label{tab:direct-vs-code}
\setlength{\tabcolsep}{4pt}
\begin{tabular}{l | l | l | cccccc}
\toprule\toprule
\textbf{Model} & \textbf{Tier} & \textbf{Cond.} & \textbf{Total} & \textbf{NA} & \textbf{FV} & \textbf{AQ} & \textbf{DM} & \textbf{CC} \\
\hline
\multirow{15}{*}{\rotatebox{90}{\textbf{GPT-5.3-Codex}}}
 & \multirow{3}{*}{Overall}
   & Direct    & 14.96 & 11.22 & 11.87 & 31.63 & 47.55 & 0.00 \\
 &           & Code      & 39.42 & 17.78 & 19.79 & 34.91 & 63.32 & 78.63 \\
 &           & $\Delta$  & \dpos{24.46} & \dpos{6.57} & \dpos{7.92} & \dpos{3.28} & \dpos{15.77} & \dpos{78.63} \\
\cmidrule(l){2-9}
 & \multirow{3}{*}{T1}
   & Direct    & 20.66 & 19.20 & 17.90 & 45.53 & --- & 0.00 \\
 &           & Code      & 47.63 & 31.20 & 27.28 & 51.99 & --- & 80.03 \\
 &           & $\Delta$  & \dpos{26.97} & \dpos{12.00} & \dpos{9.38} & \dpos{6.46} & --- & \dpos{80.03} \\
\cmidrule(l){2-9}
 & \multirow{3}{*}{T2}
   & Direct    & 6.88 & 9.40 & 8.46 & 11.51 & --- & 0.00 \\
 &           & Code      & 34.29 & 16.67 & 19.17 & 10.32 & --- & 81.37 \\
 &           & $\Delta$  & \dpos{27.42} & \dpos{7.27} & \dpos{10.72} & \dneg{1.20} & --- & \dpos{81.37} \\
\cmidrule(l){2-9}
 & \multirow{3}{*}{T3}
   & Direct    & 13.57 & 8.99 & 11.40 & 33.91 & --- & 0.00 \\
 &           & Code      & 36.97 & 12.87 & 17.27 & 39.33 & --- & 78.41 \\
 &           & $\Delta$  & \dpos{23.39} & \dpos{3.88} & \dpos{5.87} & \dpos{5.42} & --- & \dpos{78.41} \\
\cmidrule(l){2-9}
 & \multirow{3}{*}{T4}
   & Direct    & 18.73 & 7.28 & 9.71 & 29.12 & 47.55 & 0.00 \\
 &           & Code      & 38.80 & 10.40 & 15.43 & 30.13 & 63.32 & 74.72 \\
 &           & $\Delta$  & \dpos{20.07} & \dpos{3.12} & \dpos{5.72} & \dpos{1.01} & \dpos{15.77} & \dpos{74.72} \\
\hline
\multirow{15}{*}{\rotatebox{90}{\textbf{Qwen-3.5-122B-A10B}}}
 & \multirow{3}{*}{Overall}
   & Direct    & 17.63 & 14.59 & 26.56 & 24.97 & 40.24 & 0.00 \\
 &           & Code      & 37.20 & 16.14 & 21.22 & 30.29 & 63.36 & 73.67 \\
 &           & $\Delta$  & \dpos{19.57} & \dpos{1.56} & \dneg{5.34} & \dpos{5.32} & \dpos{23.12} & \dpos{73.67} \\
\cmidrule(l){2-9}
 & \multirow{3}{*}{T1}
   & Direct    & 26.36 & 26.80 & 35.41 & 43.25 & --- & 0.00 \\
 &           & Code      & 46.67 & 26.80 & 28.63 & 48.78 & --- & 82.47 \\
 &           & $\Delta$  & \dpos{20.30} & \dpos{0.00} & \dneg{6.78} & \dpos{5.53} & --- & \dpos{82.47} \\
\cmidrule(l){2-9}
 & \multirow{3}{*}{T2}
   & Direct    & 11.50 & 13.53 & 21.59 & 10.86 & --- & 0.00 \\
 &           & Code      & 32.35 & 15.53 & 19.71 & 10.78 & --- & 74.89 \\
 &           & $\Delta$  & \dpos{20.84} & \dpos{2.00} & \dneg{1.88} & \dneg{0.08} & --- & \dpos{74.89} \\
\cmidrule(l){2-9}
 & \multirow{3}{*}{T3}
   & Direct    & 14.41 & 9.63 & 24.59 & 23.42 & --- & 0.00 \\
 &           & Code      & 34.10 & 13.38 & 20.32 & 31.02 & --- & 71.68 \\
 &           & $\Delta$  & \dpos{19.69} & \dpos{3.75} & \dneg{4.27} & \dpos{7.60} & --- & \dpos{71.68} \\
\cmidrule(l){2-9}
 & \multirow{3}{*}{T4}
   & Direct    & 18.22 & 8.39 & 24.66 & 17.82 & 40.24 & 0.00 \\
 &           & Code      & 35.69 & 8.87 & 16.24 & 24.34 & 63.36 & 65.62 \\
 &           & $\Delta$  & \dpos{17.46} & \dpos{0.48} & \dneg{8.43} & \dpos{6.52} & \dpos{23.12} & \dpos{65.62} \\
\bottomrule\bottomrule
\end{tabular}
\end{table}

%% file: table/appendix_table/appendix_f_result/5_skill_vs_noskill.tex

\begin{table}[htbp]
\centering
\scriptsize
\renewcommand\arraystretch{1.3}
\caption{Code-Enabled Reasoning vs.\ Skill-Guided Code Reasoning by tier and output dimension, reported as percentages (\%). For each (model, tier) block, the three rows report \textit{Code}, \textit{Skills}, and $\Delta=$\,Skills$-$Code. \textcolor{deltapos}{\textbf{Green}} indicates Skill-Guided Code Reasoning improves over plain Code-Enabled Reasoning; \textcolor{deltaneg}{\textbf{red}} indicates degradation. Sample size $n{=}25$ per tier, $n{=}100$ overall. ``---'' indicates the dimension is not evaluated at that tier.}
\vspace{3mm}
\label{tab:code-vs-skill}
\setlength{\tabcolsep}{4pt}
\begin{tabular}{l | l | l | cccccc}
\toprule\toprule
\textbf{Model} & \textbf{Tier} & \textbf{Cond.} & \textbf{Total} & \textbf{NA} & \textbf{FV} & \textbf{AQ} & \textbf{DM} & \textbf{CC} \\
\hline
\multirow{15}{*}{\rotatebox{90}{\textbf{GPT-5.3-Codex}}}
 & \multirow{3}{*}{Overall}
   & Code      & 39.42 & 17.78 & 19.79 & 34.91 & 63.32 & 78.63 \\
 &           & Skills    & 44.94 & 17.33 & 18.60 & 60.63 & 75.64 & 77.29 \\
 &           & $\Delta$  & \dpos{5.52} & \dneg{0.45} & \dneg{1.18} & \dpos{25.72} & \dpos{12.32} & \dneg{1.34} \\
\cmidrule(l){2-9}
 & \multirow{3}{*}{T1}
   & Code      & 47.63 & 31.20 & 27.28 & 51.99 & --- & 80.03 \\
 &           & Skills    & 53.85 & 30.40 & 25.43 & 78.05 & --- & 81.52 \\
 &           & $\Delta$  & \dpos{6.22} & \dneg{0.80} & \dneg{1.85} & \dpos{26.06} & --- & \dpos{1.49} \\
\cmidrule(l){2-9}
 & \multirow{3}{*}{T2}
   & Code      & 34.29 & 16.67 & 19.17 & 10.32 & --- & 81.37 \\
 &           & Skills    & 37.45 & 17.47 & 19.89 & 30.91 & --- & 79.09 \\
 &           & $\Delta$  & \dpos{3.16} & \dpos{0.80} & \dpos{0.72} & \dpos{20.59} & --- & \dneg{2.28} \\
\cmidrule(l){2-9}
 & \multirow{3}{*}{T3}
   & Code      & 36.97 & 12.87 & 17.27 & 39.33 & --- & 78.41 \\
 &           & Skills    & 43.57 & 12.37 & 17.11 & 66.79 & --- & 78.00 \\
 &           & $\Delta$  & \dpos{6.60} & \dneg{0.49} & \dneg{0.16} & \dpos{27.46} & --- & \dneg{0.41} \\
\cmidrule(l){2-9}
 & \multirow{3}{*}{T4}
   & Code      & 38.80 & 10.40 & 15.43 & 30.13 & 63.32 & 74.72 \\
 &           & Skills    & 44.90 & 9.09 & 11.98 & 57.25 & 75.64 & 70.57 \\
 &           & $\Delta$  & \dpos{6.11} & \dneg{1.31} & \dneg{3.45} & \dpos{27.12} & \dpos{12.32} & \dneg{4.15} \\
\hline
\multirow{15}{*}{\rotatebox{90}{\textbf{Qwen-3.5-122B-A10B}}}
 & \multirow{3}{*}{Overall}
   & Code      & 37.20 & 16.14 & 21.22 & 30.29 & 63.36 & 73.67 \\
 &           & Skills    & 37.58 & 15.97 & 20.66 & 30.29 & 63.48 & 76.26 \\
 &           & $\Delta$  & \dpos{0.38} & \dneg{0.18} & \dneg{0.56} & \dpos{0.00} & \dpos{0.12} & \dpos{2.60} \\
\cmidrule(l){2-9}
 & \multirow{3}{*}{T1}
   & Code      & 46.67 & 26.80 & 28.63 & 48.78 & --- & 82.47 \\
 &           & Skills    & 46.88 & 28.00 & 28.83 & 48.35 & --- & 82.35 \\
 &           & $\Delta$  & \dpos{0.21} & \dpos{1.20} & \dpos{0.20} & \dneg{0.43} & --- & \dneg{0.12} \\
\cmidrule(l){2-9}
 & \multirow{3}{*}{T2}
   & Code      & 32.35 & 15.53 & 19.71 & 10.78 & --- & 74.89 \\
 &           & Skills    & 32.50 & 15.40 & 18.14 & 11.96 & --- & 77.02 \\
 &           & $\Delta$  & \dpos{0.16} & \dneg{0.13} & \dneg{1.57} & \dpos{1.18} & --- & \dpos{2.13} \\
\cmidrule(l){2-9}
 & \multirow{3}{*}{T3}
   & Code      & 34.10 & 13.38 & 20.32 & 31.02 & --- & 71.68 \\
 &           & Skills    & 34.40 & 11.96 & 19.28 & 30.49 & --- & 75.88 \\
 &           & $\Delta$  & \dpos{0.31} & \dneg{1.41} & \dneg{1.04} & \dneg{0.53} & --- & \dpos{4.20} \\
\cmidrule(l){2-9}
 & \multirow{3}{*}{T4}
   & Code      & 35.69 & 8.87 & 16.24 & 24.34 & 63.36 & 65.62 \\
 &           & Skills    & 36.54 & 8.51 & 16.39 & 24.50 & 63.48 & 69.80 \\
 &           & $\Delta$  & \dpos{0.85} & \dneg{0.36} & \dpos{0.15} & \dpos{0.16} & \dpos{0.12} & \dpos{4.18} \\
\bottomrule\bottomrule
\end{tabular}
\end{table}

%% file: table/appendix_table/appendix_f_result/6_skill_vs_timesage.tex
%

\definecolor{deltapos}{RGB}{0,130,60}     
\definecolor{deltaneg}{RGB}{200,30,30}    


\begin{table}[htbp]
\centering
\scriptsize
\renewcommand\arraystretch{1.3}
\caption{Skill vs.\ TimeSage comparison by tier and output dimension, reported as percentages (\%). For each (model, tier) block, the three rows report \textit{Skills}, \textit{TimeSage}, and $\Delta=$\,TimeSage$-$Skills. \textcolor{deltapos}{\textbf{Green}} indicates TimeSage improves over Skills; \textcolor{deltaneg}{\textbf{red}} indicates TimeSage degrades performance.}
\vspace{3mm}
\label{tab:skill-vs-timesage}
\setlength{\tabcolsep}{4pt}
\begin{tabular}{l | l | l | cccccc}
\toprule\toprule
\textbf{Model} & \textbf{Tier} & \textbf{Cond.} & \textbf{Total} & \textbf{NA} & \textbf{FV} & \textbf{AQ} & \textbf{DM} & \textbf{CC} \\
\hline
\multirow{15}{*}{\rotatebox{90}{\textbf{GPT-5.3-Codex}}}
 & \multirow{3}{*}{Overall}
   & Skills    & 44.94 & 17.33 & 18.60 & 60.63 & 75.64 & 77.29 \\
 & & TimeSage  & 41.50 & 19.54 & 26.46 & 40.79 & 51.61 & 76.06 \\
 & & $\Delta$  & \dneg{3.45} & \dpos{2.21} & \dpos{7.85} & \dneg{19.84} & \dneg{24.03} & \dneg{1.23} \\
\cmidrule(l){2-9}
 & \multirow{3}{*}{T1}
   & Skills    & 53.85 & 30.40 & 25.43 & 78.05 & --- & 81.52 \\
 & & TimeSage  & 47.97 & 31.20 & 29.44 & 51.55 & --- & 79.67 \\
 & & $\Delta$  & \dneg{5.88} & \dpos{0.80} & \dpos{4.01} & \dneg{26.50} & --- & \dneg{1.85} \\
\cmidrule(l){2-9}
 & \multirow{3}{*}{T2}
   & Skills    & 37.45 & 17.47 & 19.89 & 30.91 & --- & 79.09 \\
 & & TimeSage  & 39.09 & 20.20 & 23.28 & 29.72 & --- & 78.80 \\
 & & $\Delta$  & \dpos{1.63} & \dpos{2.73} & \dpos{3.38} & \dneg{1.20} & --- & \dneg{0.28} \\
\cmidrule(l){2-9}
 & \multirow{3}{*}{T3}
   & Skills    & 43.57 & 12.37 & 17.11 & 66.79 & --- & 78.00 \\
 & & TimeSage  & 39.05 & 14.45 & 26.24 & 39.04 & --- & 76.48 \\
 & & $\Delta$  & \dneg{4.51} & \dpos{2.07} & \dpos{9.13} & \dneg{27.75} & --- & \dneg{1.52} \\
\cmidrule(l){2-9}
 & \multirow{3}{*}{T4}
   & Skills    & 44.90 & 9.09 & 11.98 & 57.25 & 75.64 & 70.57 \\
 & & TimeSage  & 39.88 & 12.32 & 26.87 & 39.30 & 51.61 & 69.29 \\
 & & $\Delta$  & \dneg{5.03} & \dpos{3.22} & \dpos{14.88} & \dneg{17.95} & \dneg{24.03} & \dneg{1.27} \\
\hline
\multirow{15}{*}{\rotatebox{90}{\textbf{Qwen-3.5-122B-A10B}}}
 & \multirow{3}{*}{Overall}
   & Skills    & 37.58 & 15.97 & 20.66 & 30.29 & 63.48 & 76.26 \\
 & & TimeSage  & 38.23 & 22.54 & 23.33 & 35.04 & 35.28 & 71.15 \\
 & & $\Delta$  & \dpos{0.65} & \dpos{6.57} & \dpos{2.67} & \dpos{4.75} & \dneg{28.20} & \dneg{5.11} \\
\cmidrule(l){2-9}
 & \multirow{3}{*}{T1}
   & Skills    & 46.88 & 28.00 & 28.83 & 48.35 & --- & 82.35 \\
 & & TimeSage  & 45.56 & 38.40 & 24.96 & 43.83 & --- & 75.06 \\
 & & $\Delta$  & \dneg{1.32} & \dpos{10.40} & \dneg{3.87} & \dneg{4.53} & --- & \dneg{7.29} \\
\cmidrule(l){2-9}
 & \multirow{3}{*}{T2}
   & Skills    & 32.50 & 15.40 & 18.14 & 11.96 & --- & 77.02 \\
 & & TimeSage  & 36.79 & 20.87 & 23.07 & 26.91 & --- & 72.00 \\
 & & $\Delta$  & \dpos{4.29} & \dpos{5.47} & \dpos{4.93} & \dpos{14.95} & --- & \dneg{5.02} \\
\cmidrule(l){2-9}
 & \multirow{3}{*}{T3}
   & Skills    & 34.40 & 11.96 & 19.28 & 30.49 & --- & 75.88 \\
 & & TimeSage  & 37.05 & 16.50 & 24.60 & 35.11 & --- & 72.00 \\
 & & $\Delta$  & \dpos{2.65} & \dpos{4.53} & \dpos{5.32} & \dpos{4.61} & --- & \dneg{3.88} \\
\cmidrule(l){2-9}
 & \multirow{3}{*}{T4}
   & Skills    & 36.54 & 8.51 & 16.39 & 24.50 & 63.48 & 69.80 \\
 & & TimeSage  & 33.53 & 14.38 & 20.68 & 31.72 & 35.28 & 65.56 \\
 & & $\Delta$  & \dneg{3.01} & \dpos{5.87} & \dpos{4.30} & \dpos{7.22} & \dneg{28.20} & \dneg{4.24} \\
\bottomrule\bottomrule
\end{tabular}
\end{table}